%% file: ex_article.tex
\documentclass[review,onefignum,onetabnum]{siamart190516}

\input{ex_shared}

\ifpdf
\hypersetup{
  pdftitle={Approximate Newton policy gradient algorithms},
  pdfauthor={H. Li, S. Gupta, H. Yu, L. Ying, and I. Dhillon}
}
\fi

\begin{document}
\nolinenumbers
\maketitle

\begin{abstract}
Policy gradient algorithms have been widely applied to Markov decision processes and reinforcement
learning problems in recent years. Regularization with various entropy functions is often used to
encourage exploration and improve stability. This paper proposes an approximate Newton method for the
policy gradient algorithm with entropy regularization. In the case of Shannon entropy, the resulting
algorithm reproduces the natural policy gradient algorithm. For other entropy functions, this method
results in brand-new policy gradient algorithms. We prove that all these algorithms
enjoy Newton-type quadratic convergence and that the corresponding gradient flow converges
globally to the optimal solution. Using synthetic and industrial-scale examples, we demonstrate
that the proposed approximate Newton method typically converges in single-digit iterations, often orders
of magnitude faster than other state-of-the-art algorithms.
\end{abstract}



\section{Introduction}\label{sec:intro}

Consider an infinite-horizon Markov decision process (MDP)
\cite{bellman1957markovian,sutton2018reinforcement} $\mathcal{M} = (S, A, P, r, \gamma)$, where $S$
is a set of states of the system studied, $A$ is a set of actions made by the agent, $P$ is a
transition probability tensor with $P_{st}^{a}$ being the probability of transitioning from state
$s$ to state $t$ when taking action $a$, $r$ is a reward tensor with $r_{s}^a$ being the reward
obtained when taking action $a$ at state $s$, and $0<\gamma<1$ is a discount factor. Throughout the
paper, the state space $S$ and the action space $A$ are assumed to be finite. A policy $\pi$ is a
randomized rule of action-selection where $\pi_s^a$ denotes the probability of choosing action $a$
at state $s$. For a given policy $\pi$, the value function $v_\pi$ is defined as
\begin{equation}
  (v_\pi)_s = \E \sum_{k=0}^\infty \left(\gamma^k r_{s_k}^{a_k}\mid s_0 = s\right),
\end{equation}
which satisfies the Bellman equation:
\begin{equation}\label{eq:vanillabellman}
  (I-\gamma P_\pi)v_\pi = r_\pi,
\end{equation}
where $(P_\pi)_{st} = \sum_a\pi_s^aP_{st}^a$, $(r_\pi)_{s}=\sum_a\pi_s^ar_{s}^a$, and $I$ is the
identity operator.


In order to promote exploration and enhance stability, one often regularizes the problem with a
function $h_\pi$ such as the negative Shannon entropy $(h_\pi)_s = \sum_a \pi_s^a\log\pi_s^a$. With
the regularization $h_\pi$, the original reward $r_\pi$ is replaced with the regularized reward
$r_\pi-\tau h_\pi$ where $\tau>0$ is a regularization coefficient and \eqref{eq:vanillabellman}
becomes
\begin{equation}
  (I-\gamma P_\pi)v_\pi = r_\pi - \tau h_\pi,
\end{equation}
where we overload the notation $v_\pi$ for the regularized value function. Other continuously
differentiable entropy functions can be used as well, as we will show later. Since $\gamma < 1$ and
$P_\pi$ is a transition probability matrix, $(I-\gamma P_\pi)$ is invertible, and
\begin{equation}\label{eq:regularv}
    v_\pi = (I-\gamma P_\pi)^{-1}(r_\pi-\tau h_\pi).
\end{equation}

In a policy optimization problem, we seek a policy $\pi$ that maximizes the value function $v_\pi$. According to the theory of regularized MDPs \cite{geist2019theory}, when the regularization is strongly convex, there is a unique optimal policy $\pi^*$ such that $(v_{\pi^*})_s\geq (v_{\pi})_s$ for any policy $\pi$ and state $s$. It thus suffices to maximize $\rho^\top v_\pi$ for any positive weight vector $\rho\in\R_+^{|S|}$. Using \eqref{eq:regularv}, the problem can be stated as
\begin{equation}\label{optim}
    \underset{\pi}{\max} ~ \rho^\top(I-\gamma P_\pi)^{-1}(r_\pi-\tau h_\pi).
\end{equation}
This problem can be solved by, for example, the policy gradient (PG) method.  However, the vanilla
PG method converges quite slowly. In \cite{agarwal2020optimality}, for instance, the vanilla PG
method is shown to have the $O(T^{-1})$ convergence rate, where $T$ denotes the number of
iterations. A widely used variant of PG is the softmax policy gradient (SPG) method, where a softmax parameterization is applied before taking gradient updates, which has been shown in \cite{li2021softmax} to require $O(|S|^{2^{\Omega(\frac{1}{1-\gamma})}})$ iterations to converge for certain MDPs without regularization. For the PG method with entropy regularization and some of its variants, the convergence
rate can be improved to $O(e^{-cT})$, i.e., linear convergence \cite{mei2020global}, which can
still be slow since the constant $c$ in the linear convergence rate $O(e^{-cT})$ is in general close to $0$.  It is also demonstrated in numerical examples
that these algorithms with linear rates can experience slow convergence.  For example,
in the example in \cite{zhan2021policy}, thousands of iterations are needed for the algorithm to
converge, even though the model is relatively small and sparse. Therefore, there is a clear need for
designing new methods with faster convergence and one idea is to take the geometry of the problem
into consideration. The Newton method, for example, preconditions the gradient with the Hessian
matrix and obtains second-order local convergence. Since the exact Hessian matrix is usually too
computationally expensive to obtain, the approximate Newton methods (including quasi-Newton methods), which use structurally simpler approximations of the Hessian instead, are more widely adopted in generic optimization problems and are known to enjoy superlinear convergence \cite{rodomanov2021new, rodomanov2021rates}.

\subsection{Contributions}
In this paper, we investigate the approximate Newton approach for solving \eqref{optim}. The main
contributions of this paper are the following.
\begin{itemize}
\item First, we present a unified approximate Newton method for the policy optimization problem.  The
  main observation is to decompose the Hessian as a sum of a diagonal matrix and a remainder that vanishes at the optimal solution. This inspires us to use only the diagonal matrix in the approximate Newton method. As a result, the proposed method not only leverages the second-order information but also enjoys low computational cost due to the diagonal structure of the preconditioner used. When the negative Shannon entropy is used,
  this method reproduces the natural policy gradient (NPG) algorithm. For other forms of entropy regularization, this method results in brand-new policy gradient algorithms.
  
\item Second, we analyze the convergence property of the proposed approximate Newton algorithms and
  demonstrate local quadratic convergence both theoretically and numerically. By leveraging the framework of Newton-type methods (see \cite{dennis1974characterization} for example), we provide a simple and straightforward proof for quadratic convergence near the optimal policy. In the numerical tests, we verify that the proposed method leads to fast quadratic convergence even under small regularization and large discount rates (close to 1). Even for industrial-size problems with hundreds of thousands of states, the approximate Newton method converges in single-digit iterations and within a few minutes on a regular laptop. We also prove the global convergence of the approximate Newton gradient flow to the optimal solutions. 
\end{itemize}

\subsection{Background and related work}
A major workhorse behind the recent success of reinforcement learning (RL) is the large family of
policy gradient (PG) methods \cite{williams1992simple, sutton1999policy}, for
example, the natural policy gradient (NPG) method \cite{kakade2001natural}, the actor-critic method \cite{konda2000actor}, the asynchronous advantage actor-critic (A3C) method
\cite{mnih2016asynchronous}, the deterministic policy gradient (DPG) method
\cite{silver2014deterministic}, the trust region policy optimization (TRPO)
\cite{schulman2015trust}, the generalized advantage estimation (GAE) \cite{schulman2015high}, and
proximal policy optimization (PPO) \cite{schulman2017proximal}, to mention but a few. The NPG method is known to be drastically faster than the original PG method because the policy gradient in NPG is preconditioned by the Fisher information (an approximation of the Hessian of the KL-divergence) matrix in order to fit the problem geometry better. This idea is extended in TRPO and PPO where the problem geometry is taken into consideration via trust region constraints (in terms of KL-divergence) and a clipping function of the relative ratio of policies in the objective function, respectively. These implicit ways (in the sense that they do not adjust the gradient by an explicit preconditioner) of adjusting the policy gradient are in essence similar to the mirror descent (MD) method \cite{nemirovskij1983problem} in generic optimization problems.

This similarity in addressing the inherent geometry of the problem is noticed by a line of recent
work including \cite{neu2017unified, geist2019theory, shani2020adaptive, tomar2020mirror, lan2021policy}, and the analysis techniques in MD methods have been adapted to the PG setting. The connection was first built explicitly in \cite{neu2017unified}. The authors consider a linear program formulation where the objective function is the average reward and the domain is the set of stationary state-action distributions, in which case the TRPO method can be viewed as an approximate mirror descent method and the A3C method as an MD method for the dual-averaging \cite{nesterov2009primal} objective. As a complement, \cite{geist2019theory} considers an actor-critic type method where the policy is updated via either a regularized greedy step or an MD step, and the value function is updated by a regularized Bellman operator, which also includes TRPO as a special case, and error propagation analysis is provided. In \cite{shani2020adaptive}, an adaptive scaling that naturally arises in the policy gradient is applied to the proximity term of the MD formulation, and the sublinear convergence result is proved with a properly decreasing learning rate. In \cite{tomar2020mirror}, the application to the non-tabular setting is enabled by parameterizing the policy and applying MD to the policy parameters, and the corresponding sublinear convergence result is presented. 

Regularization, a strategy that considers the modified objective function with an additional penalty term on the policy, is another crucial component in the development of PG-type methods. Intuitively, regularization is able to encourage exploration in the policy iteration process and thus avoid local minima. It is also suggested \cite{ahmed2019understanding} that regularization makes the optimization landscape smoother and thus enables possibly faster convergence. Linear convergence results are then established for regularized PG and NPG methods \cite{agarwal2020optimality, mei2020global, cen2020fast}. In these relatively earlier works \cite{agarwal2020optimality, mei2020global, cen2020fast}, the regularization usually takes the form of (negative) entropy or relative entropy. In the more recent work \cite{lan2021policy} and \cite{zhan2021policy} that follow the MD type methods, the regularization is extended to general convex functions with the resulting Bregman divergences different from the KL-divergence and linear convergence is guaranteed as well. 

However, most of these algorithms are of either sublinear or linear convergence except the entropy
regularized NPG with full step length (which is a special case of the approximate Newton method we
propose), and even the linear convergence rate $O(e^{-cT})$ can be slow since $c$ can be close to
zero. This motivates us to invent the approximate Newton policy gradient method to be introduced in \cref{sec:Newton}.

\section{Approximate Newton method}\label{sec:Newton}
\subsection{Approximate Newton method and entropy regularized natural policy gradient}
This section derives the approximate Newton method for the entropy regularized policy optimization
problems. The idea is to approximate the Hessian with a simpler matrix whose inverse is easy to compute.
We start with the negative Shannon entropy $(h_\pi)_s = \sum_a \pi_s^a\log\pi_s^a$.

In what follows, it is assumed that $\pi^*$ is the optimizer of the problem stated in \eqref{optim}.  By
introducing $Z_\pi := I-\gamma P_\pi$, the objective function can be written as
\begin{equation}
  E(\pi) \equiv \rho^\top(I-\gamma P_\pi)^{-1}(r_\pi-\tau h_\pi) = \rho^\top Z_\pi^{-1}(r_\pi-\tau h_\pi) = w_\pi^\top(r_\pi-\tau h_\pi), 
\end{equation}
where $w_\pi := Z_\pi^{-\top}\rho$. 

Let us first outline the main idea of the approximate Newton method. The gradient $\nabla_\pi E$ in $\R^{|S||A|}$ of $E(\pi)$ has entries given by
\begin{equation}
  \frac{\partial E}{\partial \pi_s^a} = (r_s^a - \tau (\log\pi_s^a +1) - \left[(I-\gamma P^a)v_\pi\right]_s + c_s)(w_\pi)_s,
\end{equation}
where $c_s$ is a multiplier associated with the constraint $\sum_a\pi_s^a = 1$ that depends on $s$. Our key observation is to decompose the Hessian matrix $D^2E(\pi)$ in $\R^{|S||A|\times |S||A|}$ into two parts
\begin{equation}\label{eq:decomp}
D^2E(\pi) = \Lambda(\pi) + \Delta(\pi),
\end{equation}
where $\Lambda(\pi)$ is a {\em diagonal} matrix given by
$\Lambda_{(sa),(tb)}=-\tau\delta_{\{(sa),(tb)\}}\frac{(w_\pi)_s}{\pi_s^a}$ and $\Delta(\pi)$ is a
remainder that {\em vanishes} at $\pi = \pi^*$, i.e., $\Delta(\pi)=O(\|\pi-\pi^*\|)$ (shown in \Cref{thm: approxnewton}). We emphasize that $\Lambda(\pi)$ is in general not the diagonal part of the Hessian matrix $D^2E(\pi)$, but a diagonal approximation to it. With this decomposition, we can approximate the Hessian matrix $D^2E(\pi)$
by $\Lambda(\pi)$ and obtain the following {\em approximate Newton flow}:
\[
\begin{aligned}
  \frac{\mathrm{d}\pi_s^a}{\mathrm{d} t} &=-(\Lambda^{-1}\nabla_\pi E)_{sa}= -(\Lambda_{(sa),(sa)})^{-1}\frac{\partial E}{\partial \pi_s^a}\\
  &= \pi_s^a(r_s^a-\tau(\log\pi_s^a+1)-[(I-\gamma P^a)v_\pi]_s + c_s)/\tau,
\end{aligned}
\]
By introducing the parameterization $\theta_s^a = \log \pi_s^a$ and discretizing in time with learning rate $\eta$, we arrive at
\[
\theta_s^a \leftarrow \eta(r_s^a-\tau-[(I-\gamma P^a)v_\pi]_s + c_s)/\tau + (1-\eta)\theta_s^a.
\]
Writing this update back in terms of $\pi_s^a$ leads to the following update rule
\[
\pi_s^a \propto (\pi_s^a)^{1-\eta}\exp(\eta(r_s^a+(\gamma
  P^av_\pi)_s)/\tau),
\]
which coincides with the NPG scheme with entropy regularization. This result is summarized in the following theorem with the proof given in \cref{sec:approxnewton}. 

\begin{theorem}\label{thm: approxnewton}
  Let $h_\pi \in \R^{|S|}$ be the negative Shannon entropy $(h_\pi)_s=\sum_a\pi_s^a\log\pi_s^a$.

  (a) There exists a diagonal approximation $\Lambda(\pi)$ of the Hessian matrix $D^2E(\pi)$ given by $\Lambda_{(sa),(tb)}=-\tau\delta_{\{(sa),(tb)\}}\frac{(w_\pi)_s}{\pi_s^a}$ such that
  \begin{equation}\label{eq:approxH}
      \Lambda(\pi)-D^2E(\pi) = O(\|\pi-\pi^*\|).
  \end{equation}

  (b) The approximate Newton flow from $\Lambda(\pi)$ is
  \begin{equation}\label{eq:QNflow}
    \frac{\mathrm{d}\pi_s^a}{\mathrm{d} t} = \pi_s^a(r_s^a-\tau(\log\pi_s^a+1)-[(I-\gamma P^a)v_\pi]_s+c_s)/\tau.
  \end{equation}
  With a learning rate $\eta$, the gradient update is
  \begin{equation}\label{eq:NPG}
    \pi_s^a \leftarrow \frac{(\pi_s^a)^{1-\eta}\exp(\eta(r_s^a+(\gamma
      P^av_\pi)_s)/\tau)}{\sum_a(\pi_s^a)^{1-\eta}\exp(\eta(r_s^a+(\gamma P^av_\pi)_s)/\tau)}.
  \end{equation}
\end{theorem}

\begin{remark}
  The policy update scheme \eqref{eq:NPG} is the same as the entropy regularized natural policy
  gradient scheme in, for example, \cite{cen2020fast}.
\end{remark}

\paragraph{Historical note}
The natural gradient methods (including the NPG method) were traditionally developed as a way of
implementing the vanilla gradient descent method with an intrinsic metric that is invariant to the
choice of parameters (Cf. \cite{martens2020new}), and entropy regularization was originally
motivated as a way of encouraging exploration and avoid the suboptimality caused by greedy solvers
(Cf. \cite{neu2017unified}). In this regard, it was more or less a coincidence that the algorithm
combining the two methods -- the regularized NPG obtains a fast quadratic convergence
(Cf. \cite{cen2020fast}). The reason behind this coincidence is that the preconditioner used in the
natural gradient method in fact approximates the second-order derivatives introduced by the entropy
regularization in this case, though the fisher information matrix was not designed to approximate
any second-order information in the classical natural gradient literature.

\subsection{Extension to other entropy functions}
\Cref{thm: approxnewton} can be extended to more general entropy functions. It yields brand-new
algorithms with quadratic convergence. Here we consider the entropy functions of the form
\begin{equation}\label{eq:generaletp}
(h_\pi)_s = \sum_a\phi\left(\frac{\pi_s^a}{\mu_s^a}\right)\mu_s^a,
\end{equation}
where $\phi$ is convex on $(0, +\infty)$ and $\phi(1) = 0$, and $\mu_s$ is a prior distribution over $A$ such that $\mu_s^a>0$. The term $(h_\pi)_s$ is also called the ``$f$-divergence'' between $\pi_s$ and $\mu_s$ \cite{renyi1961measures,ali1966general}. If there is no prior knowledge of the policy, one can use the uniform prior, i.e., $\mu_s^a=1/|A|$ for all $a$. 
We further assume that $\phi$ is twice continuously differentiable and strongly convex and that $\phi'(x)\rightarrow-\infty$ as $x\rightarrow 0$. Here are some examples:
\begin{itemize}
\item When $\phi(x) = x\log x$, $(h_\pi)_s = \sum_a
  \left(\frac{\pi_s^a}{\mu_s^a}\log\frac{\pi_s^a}{\mu_s^a}\right)\mu_s^a
  =\sum_a\pi_s^a\log\frac{\pi_s^a}{\mu_s^a}$. When the uniform prior is used, we recover the
  negative Shannon entropy regularization $\sum_a \pi_s^a\log\pi_s^a$ used in \Cref{thm:
    approxnewton} after omitting the constant $\log\frac{1}{|A|}$.
\item When $\phi(x) = \frac{4}{1-\alpha^2}(1-x^{\frac{1+\alpha}{2}})~
  (\alpha<1)$, we obtain the $\alpha$-divergence:
  \begin{equation}\label{eq:alpha}
    (h_\pi)_s = \frac{4}{1-\alpha^2}-\frac{4}{1-\alpha^2}\sum_a\mu_s^a\left(\pi_s^a/\mu_s^a\right)^{\frac{1+\alpha}{2}}.
  \end{equation}
  In particular, when $\alpha=0$ we obtain the Hellinger divergence
  $(h_\pi)_s=2-2\sum_a\sqrt{\mu_s^a\pi_s^a}$ after dividing by $2$. When
  $\alpha\rightarrow-1$ we obtain the reverse-KL divergence $(h_\pi)_s =   \sum_a\mu_s^a\log\frac{\mu_s^a}{\pi_s^a}$. Also, when $\alpha\rightarrow1$, we obtain the KL-divergence $(h_\pi)_s=\sum_a\pi_s^a\log\frac{\pi_s^a}{\mu_s^a}$, though the limit of $\phi(x)$ does not exist when $\alpha\rightarrow1$. 
\end{itemize}

In the following theorem, we extend the approximate Newton method in \Cref{thm: approxnewton} to the entropy functions described above. The proof of this theorem can be found in \cref{sec:otheretp}. 

\begin{theorem}\label{thm: otheretp}
  Assume that $\pi^*$ is the optimizer of \eqref{optim} where $h_\pi$ is the entropy function defined in \eqref{eq:generaletp}.

  (a) The Hessian matrix $D^2E(\pi)$ can be approximated by a diagonal matrix $\Lambda(\pi)$ given by 
  \begin{equation}\label{eq:generalapproxhess}
    \Lambda_{(sa),(tb)} = -\tau\delta_{\{(sa),(tb)\}}\frac{(w_\pi)_s\phi''(\pi_s^a/\mu_s^a)}{\mu_s^a}
  \end{equation} 
  near $\pi^*$ such that $\Lambda(\pi)-D^2E(\pi) = O(\|\pi-\pi^*\|)$.    
  
  (b) The approximate Newton flow from $\Lambda$ is
  \begin{equation}\label{eq:qNflowgen}
  \frac{\mathrm{d}\pi_s^a}{\mathrm{d} t} = \mu_s^a(\phi''(\pi_s^a/\mu_s^a))^{-1}(r_s^a- \tau \phi'(\pi_s^a/\mu_s^a)-[(I-\gamma P^a)v_\pi]_s+c_s)/\tau.
  \end{equation}
  With parameterization $\theta_s^a = \phi'(\pi_s^a/\mu_s^a)$, the approximate Newton method from $\Lambda(\pi)$ can be expressed as:
  \begin{equation}\label{eq:quasiNTupdate}
    \theta_s^a \leftarrow \eta(r_s^a-[(I-\gamma P^a)v_\pi]_s + c_s)/\tau + (1-\eta)\theta_s^a.
  \end{equation}
  where where $0<\eta\leq1$ is the learning rate and $c_s$ is a multiplier introduced by the constraint $\sum_a\pi_s^a = 1$.

\end{theorem}

For particular choices of $\phi$, the corresponding approximate Newton update scheme can be obtained directly by plugging $\phi$ into \eqref{eq:quasiNTupdate}. 
\begin{itemize}
    \item For the case $\phi(x) = x\log x$ and $(h_\pi)_s=\sum_a\pi_s^a\log\frac{\pi_s^a}{\mu_s^a}$,
one can solve the multipliers $c_s$ explicitly as in \Cref{thm: approxnewton} and obtain the NPG method with prior distribution $\mu$:
\begin{equation}
  \pi_s^a \leftarrow \frac{(\mu_s^a)^\eta(\pi_s^a)^{1-\eta}\exp(\eta(r_s^a+(\gamma P^av_\pi)_s)/\tau)}{\sum_a(\mu_s^a)^\eta(\pi_s^a)^{1-\eta}\exp(\eta(r_s^a+(\gamma P^av_\pi)_s)/\tau)}.
\end{equation}
    \item For the case $\phi(x) = \frac{4}{1-\alpha^2}(1-x^{\frac{1+\alpha}{2}})~(\alpha<1)$ and $h_\pi$ given by the $\alpha$-divergence \eqref{eq:alpha}, we have $\theta_s^a = -\frac{2}{1-\alpha}(\pi_s^a/\mu_s^a)^{\frac{\alpha-1}{2}}$, thus by \eqref{eq:quasiNTupdate} the update scheme is 
\begin{equation}\label{eq:alphaupdate}
    \pi_s^a \leftarrow \mu_s^a\left((1-\eta)(\pi_s^a/\mu_s^a)^{\frac{\alpha-1}{2}} + \frac{\alpha-1}{2}\eta(r_s^a-[(I-\gamma P^a)v_\pi]_s+c_s)/\tau\right)^{\frac{2}{\alpha-1}}.
\end{equation}
\end{itemize}

The remaining problem in the update schemes is the determination of the multipliers $c_s$,
since in general they cannot be solved explicitly as in the case of the negative Shannon entropy (Cf. \Cref{thm: approxnewton}). Since $\phi$ is strongly convex, we know that $\phi'$ is strictly increasing, and thus $-\phi'$ is a strictly decreasing function mapping from $(0, +\infty)$ to $(-\sup\phi', +\infty)$ since $\underset{x\rightarrow0+0}{\lim}\phi'(x)=-\infty$. Let $\psi:=(-\phi')^{-1}$, then $\psi: (-\sup\phi', +\infty)\rightarrow (0, +\infty) $ is a strictly decreasing function that satisfies $\underset{x\rightarrow-\sup\phi'+0}{\lim}\psi(x)=+\infty$ and $\underset{x\rightarrow+\infty}{\lim}\psi(x)=0$. From \eqref{eq:quasiNTupdate}, the equation of the multiplier $c_s$ corresponding to $\sum_a\pi_s^a=1$ is:
\begin{equation}\label{eq:eqmultorig}
    \sum_a\mu_s^a\psi\left(-\frac{\eta}{\tau}c_s-(1-\eta)\phi'\left(\pi_s^a/\mu_s^a\right) -\frac{\eta}{\tau}(r_s^a-[(I-\gamma P^a)v_\pi]_s)\right)=1,
\end{equation}
or equivalently,
\begin{equation}\label{eq:eqmult}
    \sum_a\mu_s^a\psi\left(\tilde{c}_s+x_a\right)=1,
\end{equation}
where 
\begin{equation}
    \label{eq:deftildecxa}
    \tilde{c}_s=-\frac{\eta}{\tau}c_s, \quad
    x_a = -(1-\eta)\phi'\left(\pi_s^a/\mu_s^a\right) -\frac{\eta}{\tau}(r_s^a-[(I-\gamma P^a)v_\pi]_s).
\end{equation}
We claim that the determination of $\tilde{c}_s$ in equation \eqref{eq:eqmult} (and thus the determination of $c_s$) can be done in a similar way as in \cite{ying2020mirror} based on the following lemma. The proof of this lemma can be found in \cref{sec:mult}. 

\begin{lemma}\label{lemma:multipliergen}
  Let $L\in\R\cup\{-\infty\}$. Assume that $\psi: (L, +\infty)\rightarrow (0, +\infty) $ is a strictly decreasing function that satisfies $\underset{x\rightarrow L+0}{\lim}\psi(x)=+\infty$ and $\underset{x\rightarrow+\infty}{\lim}\psi(x)=0$ and assume also that $\mu_i>0$, then for any $x_1, x_2, \ldots, x_k$, there is a unique solution to the equation:
  \eq{\label{eq:lemmagen}
    \mu_1\psi(x+x_1) + \cdots + \mu_k\psi(x+x_k) = 1.
  }
  Moreover, the solution is on the interval
  \begin{equation}\label{bisect_intervalgen}
  \left[\max\left\{L-\underset{1\leq i\leq k}{\min}x_i, \underset{1\leq i\leq k}{\min}\left\{\psi^{-1}\left(\frac{1}{k\mu_i}\right) - x_i\right\}\right\}, \underset{1\leq i\leq k}{\max}\left\{\psi^{-1}\left(\frac{1}{k\mu_i}\right) - x_i\right\}\right].
  \end{equation}
\end{lemma}

Leveraging \Cref{lemma:multipliergen} and the monotonicity of the function $\mu_1\psi(x+x_1) + \cdots + \mu_k\psi(x+x_k) - 1$, many of the established numerical methods (e.g. bisection) for nonlinear equations can be applied to determine the solution for \eqref{eq:lemmagen}. This routine can be used to find $\tilde{c}_s$ in \eqref{eq:eqmult} and thus the multipliers $c_s$ in \eqref{eq:eqmultorig} as stated in \Cref{prop:multiplier} whose proof is given in \cref{sec:multsolve}. 

\begin{prop}
\label{prop:multiplier}
  The multipliers $c_s$ in the update scheme \eqref{eq:quasiNTupdate} can be determined uniquely
  such that the updated policy $\pi$ satisfies $\pi_s^a\geq0$ and $\sum_a\pi_s^a =1$.
\end{prop}

When the $\alpha$-divergence is used, we have $\phi=\frac{4}{1-\alpha^2}(1-x^{\frac{1+\alpha}{2}})$ and $\phi'(x) = \frac{2}{\alpha-1}x^{\frac{\alpha-1}{2}}$, then $L=-\sup\phi'=0$ and $\psi(x) = (-\phi')^{-1}(x) = (\phi')^{-1}(-x) = (\frac{1-\alpha}{2}x)^{\frac{2}{\alpha-1}}$. The algorithm proposed in this section is summarized in \Cref{alg:quasi} below. 
\begin{algorithm}[htbp]
    \caption{Approximate Newton method for the regularized MDP} 
    \label{alg:quasi}
    \begin{algorithmic}[1]
        \REQUIRE the MDP model $\mathcal{M} = (S, A, P, r, \gamma)$, initial policy $\pi_{\text{init}}$, convergence threshold $\epsilon_{\text{tol}}$,
        regularization coefficient $\tau$, learning rate $\eta$, the regularization type (KL or $\alpha$-divergence).
        \STATE Initialize the policy $\pi = \pi_{\text{init}}$.
        \STATE Set $\xi = 1 + \epsilon_{\text{tol}}$ and $k = |A|$.

        \WHILE {$\xi > \epsilon_{\text{tol}}$}
        
        \STATE Calculate the regularization term $h_\pi$ by $(h_\pi)_s = \sum_a\mu_s^a\phi(\pi_s^a/\mu_s^a)$. 
        \STATE Calculate $P_\pi$ and $r_\pi$ by $(P_\pi)_{st} = \sum_a\pi_s^aP_{st}^a, (r_\pi)_{s} = \sum_a\pi_s^ar_{s}^a$.
        \STATE Calculate $v_\pi$ by \eqref{eq:regularv}, i.e., $v_\pi = (I-\gamma P_\pi)^{-1}(r_\pi-\tau h_\pi)$.

        \IF {the KL divergence is used}
        \STATE $(\pi_{\text{new}})_s^a \leftarrow \frac{(\mu_s^a)^\eta(\pi_s^a)^{1-\eta}\exp(\eta(r_s^a+(\gamma P^av_\pi)_s)/\tau)}{\sum_a(\mu_s^a)^\eta(\pi_s^a)^{1-\eta}\exp(\eta(r_s^a+(\gamma P^av_\pi)_s)/\tau)}$ for $a = 1, 2, \ldots, |A|, ~ s = 1, 2, \ldots, |S|$.
        \ENDIF

        \IF {the $\alpha$-divergence is used}

        \FOR {$s = 1, 2, \ldots, |S|$}

        \STATE Set $L=0$ and $\psi(x) = (\frac{1-\alpha}{2}x)^{\frac{2}{\alpha-1}}$
        \STATE Calculate $x_a = -(1-\eta)\phi'\left(\pi_s^a/\mu_s^a\right) -\frac{\eta}{\tau}(r_s^a-[(I-\gamma P^a)v_\pi]_s), ~ a = 1, \ldots, |A|$.        

        \STATE Solve for $\tilde{c}_s=-\frac{\eta}{\tau}c_s$ with the bisection method on the interval described in \eqref{bisect_intervalgen}. 
        \STATE Update $(\pi_{\text{new}})_s^a \leftarrow \mu_s^a \psi(\tilde{c}_s+x_a)$ for $a = 1, 2, \ldots, |A|$.
        \ENDFOR
        \ENDIF

        \STATE $\xi = \norm{\pi_{\text{new}}-\pi}_F / \norm{\pi}_F$.
        \STATE $\pi = \pi_{\text{new}}$

        \ENDWHILE
    \end{algorithmic}
    \vspace{0.5em}
\end{algorithm}
\subsection{Convergence of the approximate Newton gradient flow}\label{sec:flowconv}
Recall from \eqref{eq:qNflowgen} that the approximate Newton gradient flow with the general entropy functions is
\[
\frac{\mathrm{d}\pi_s^a}{\mathrm{d} t} = \mu_s^a(\phi''(\pi_s^a/\mu_s^a))^{-1}(r_s^a- \tau \phi'(\pi_s^a/\mu_s^a)-[(I-\gamma P^a)v_\pi]_s+c_s)/\tau,
\]
from which we can obtain the dynamics of the objective function $E$:
\begin{equation}
\begin{aligned}
\frac{\mathrm{d}E(\pi)}{\mathrm{d} t} &= \sum_{sa} \frac{\partial E}{\partial \pi_s^a}\frac{\mathrm{d}\pi_s^a}{\mathrm{d} t}\\
&= \sum_{sa}\bigg[(r_s^a - \tau \phi'(\pi_s^a/\mu_s^a) - \left[(I-\gamma P^a)v_\pi\right]_s + c_s)(w_\pi)_s\\
&\;\;\; \cdot\mu_s^a(\phi''(\pi_s^a/\mu_s^a))^{-1}(r_s^a- \tau \phi'(\pi_s^a/\mu_s^a)-[(I-\gamma P^a)v_\pi]_s+c_s)/\tau \bigg]\\
&= \sum_{sa}\mu_s^a(\tau\phi''(\pi_s^a/\mu_s^a))^{-1}(r_s^a- \tau \phi'(\pi_s^a/\mu_s^a)-[(I-\gamma P^a)v_\pi]_s+c_s)^2(w_\pi)_s\\
&\geq 0,
\end{aligned}
\end{equation}
where we have used the gradient
\begin{equation}
    \frac{\partial E}{\partial \pi_s^a} = (r_s^a - \tau \phi'(\pi_s^a/\mu_s^a) - \left[(I-\gamma P^a)v_\pi\right]_s + c_s)(w_\pi)_s.
\end{equation}
As a result, we have shown that $\frac{\mathrm{d}E(\pi)}{\mathrm{d} t}\geq0$. Since $E(\pi)$ is upper-bounded by $\rho^\top v_{\pi^*}$, $E(\pi)$ converges. With a closer look, the following theorem states that the limiting policy is exactly the optimal policy $\pi^*$ and the proof is given in \cref{sec:flowconvpf}.
\begin{theorem}\label{thm:flowconv}
The approximate Newton flow \eqref{eq:qNflowgen} converges globally to the optimal policy $\pi^*$. 
\end{theorem}

\section{Quadratic Convergence}\label{sec:analysis}
In this section, we study the quadratic convergence of the approximate Newton method at the learning rate
$\eta=1$, which corresponds to the step size used in the Newton method. Our analysis is inspired by the results in \cite{dennis1974characterization,
  wang2021hessian}. The following theorem states the second-order convergence when $\eta=1$, with
the proof given in \cref{sec:quadconvpf}. For the simplicity of notations, we let $f(\pi)_{sa} = -(r_s^a-((I-\gamma P^a)v_\pi)_s)$, which is the additive inverse of the advantage function.

\begin{theorem}\label{thm:quadconv}
Let
\[
\Phi(\pi) = \tau\sum_{sa}\mu_s^a\phi\left(\frac{\pi_s^a}{\mu_s^a}\right),
\]
where $\phi$ is twice continuously differentiable and strongly convex and that $\phi'(x)\rightarrow-\infty$ as $x\rightarrow 0$. 
Denote the $k$-th policy obtained in \Cref{alg:quasi} by $\pi\sps{k}$. For $\eta=1$, the update scheme in \Cref{alg:quasi} can be summarized as
  \begin{equation}\label{eq:Phiupdate}
  \grad\Phi(\pi\sps{k+1})-\grad\Phi(\pi\sps{k}) = -\left(f(\pi\sps{k}) + \grad\Phi(\pi\sps{k}) -
  B^\top c(\pi\sps{k})\right),
  \end{equation}
where $f(\pi)_{sa} = -(r_s^a-((I-\gamma P^a)v_\pi)_s)$ and we denote by $B$ the $|S|$-by-$(|S|\times|A|)$ matrix such that $B_{ij} = 1$ for $|A|(i-1)+1\leq j\leq|A|i$ and $B_{ij}=0$ otherwise. 
Then $\pi\sps{k}$ enjoys a quadratic local convergence to $\pi^*$, i.e., $\lim_{k\rightarrow\infty}\pi\sps{k}=\pi^*$ and
  \begin{equation}\label{eq:quadconv}
    \norm{\pi\sps{k+1}-\pi^*} \leq C\norm{\pi\sps{k}-\pi^*}^2,
  \end{equation}
for some constant $C$, given that the initial policy $\pi\sps{0}$ is sufficiently close to $\pi^*$.
\end{theorem}
\begin{remark}
  It is clear that the quadratic convergence also occurs if $\pi\sps{k}$ is in a sufficiently small neighborhood of $\pi^*$ for some $k\geq1$ even if $\pi\sps{0}$ is not. The precise description of this small neighborhood is provided in the proof (see \Cref{sec:quadconvpf}). For a special case of this result, where $\phi(x)=x\log x$ and $\mu_s^a=1/|A|$, the algorithm is reduced to the entropy regularized NPG. A similar local convergence result for this special case has been obtained in \cite{cen2020fast}, where the proof leverages the particular structure of Shannon entropy. 
\end{remark}

\paragraph{Connection with mirror descent}
The approximate Newton algorithm \eqref{eq:Phiupdate} for $\eta=1$ has a deep connection with mirror
descent. The vanilla mirror descent of $-E(\pi)$ with a learning rate $\beta$ and the Bregman
divergence associated with $\Phi$ is given by
\eqs{
  \pi\sps{k+1} &= \underset{\pi}{\arg\min}\{-E(\pi\sps{k})-\grad E(\pi\sps{k})(\pi-\pi\sps{k}) + \frac{1}{\beta}(\Phi(\pi)-\Phi(\pi\sps{k})-\grad\Phi(\pi\sps{k})(\pi-\pi\sps{k}))\}\\
    &= \underset{\pi}{\arg\min}\{(\diag(w_{\pi\sps{k}})\otimes I_{|A|})(f(\pi\sps{k})+\grad\Phi(\pi\sps{k}))(\pi-\pi\sps{k}) + \frac{1}{\beta}(\Phi(\pi)-\grad\Phi(\pi\sps{k})\pi)\},
}
where $\diag(w_{\pi\sps{k}})$ is the diagonal matrix with the diagonal equal to
$w_{\pi\sps{k}} := (I-\gamma P_{\pi\sps{k}}^\top)^{-1}\rho$, $\otimes$ denotes the Kronecker product,
and $I_{|A|}$ denotes the $|A|$ by $|A|$ identity matrix. In the last equality, the terms
independent of $\pi$ are dropped and the multiplier term in $\grad E$ is canceled out using $B\pi =
B\pi\sps{k} = \mathbf{1}_{|S|}$. The first-order stationary condition of this minimization problem
reads
\eq{
    \grad\Phi(\pi\sps{k+1}) - \grad\Phi(\pi\sps{k}) = -\beta(\diag(w_{\pi\sps{k}})\otimes I_{|A|})(f(\pi\sps{k}) + \grad\Phi(\pi\sps{k}) - B^\top c(\pi\sps{k})).
}
This suggests that \eqref{eq:Phiupdate} can be reinterpreted as an {\em accelerated} mirror descent method with {\em adaptive} learning rates $\beta_s \equiv 1/(w_{\pi\sps{k}})_s$ that depend on the state $s$ and the current iterate $\pi\sps{k}$. Observation of the connection between mirror descent and the natural gradient method (which is similar with the approximate Newton method in this paper when the Shannon entropy is used) is given in \cite{raskutti2015information, gunasekar2021mirrorless}. 

In \cite{zhan2021policy}, a variant of mirror descent is proposed based on an implicit update scheme 
\eq{
    (\grad\Phi(\pi\sps{k+1}))_{sa} - (\grad\Phi(\pi\sps{k}))_{sa} = -\beta' \left(f(\pi\sps{k})_{sa} + \grad\Phi(\pi\sps{k+1})_{sa} - (c(\pi\sps{k}))_s\right),
}
with a state independent learning rate $\beta'$. In the next section, we will compare this variant with our approximate Newton method \eqref{eq:Phiupdate} and show that the approximate Newton method converges orders of magnitudes faster than the ones in \cite{zhan2021policy}.

\section{Numerical results}\label{sec:numerical}

\subsection{Experiment I}
\label{sec:exptoy}

We first test the approximate Newton methods derived in \cref{sec:Newton} on the model in
\cite{zhan2021policy}. For the sake of completeness, we include the description of the model here. The MDP considered has a state space $S$ of size $200$ and an action space $A$ of size $50$. For each state $t$ and action $a$, a subset $S_{t}^a$ of $S$ is uniformly randomly chosen such that $|S_t^a| = 20$, and $P_{tt'}^a = 1/20$ for any $t'\in S_t^a$. The reward is given by $r_s^a = U_s^aU_s$, where $U_{s}^a$ and $U_s$ are independently uniformly chosen on $[0,1]$. The discount rate $\gamma$ is set as $0.99$ and the regularization coefficient $\tau=0.001$.

\begin{figure}[ht]
  \centering
  \subfigure[Relative change of the policy using \Cref{alg:quasi} and methods from \cite{zhan2021policy}.\label{fig:toy1}]{
    \includegraphics[width=0.3\textwidth]{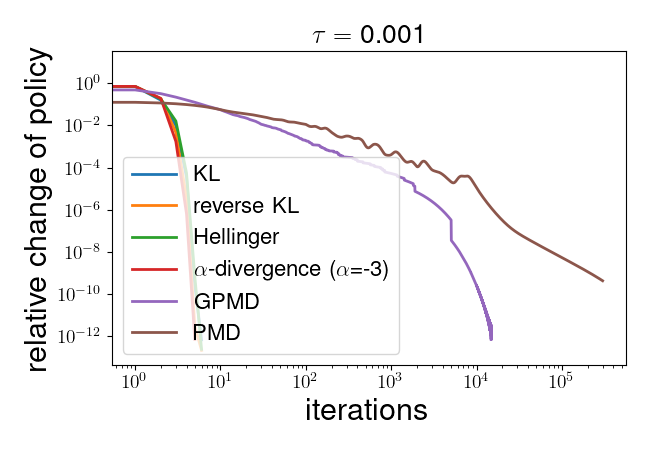} }
  \hspace{0.7em}
  \subfigure[The policy error in the process of training using KL-divergence. \label{fig:model1_KL}]{
    \includegraphics[width=0.3\textwidth]{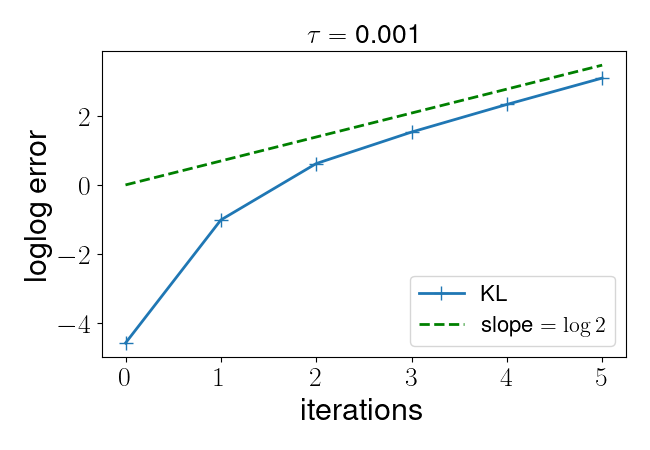} }
  \hspace{0.7em}
  \subfigure[The policy error in the process of training using reverse KL-divergence. \label{fig:model1_rKL}]{
    \includegraphics[width=0.3\textwidth]{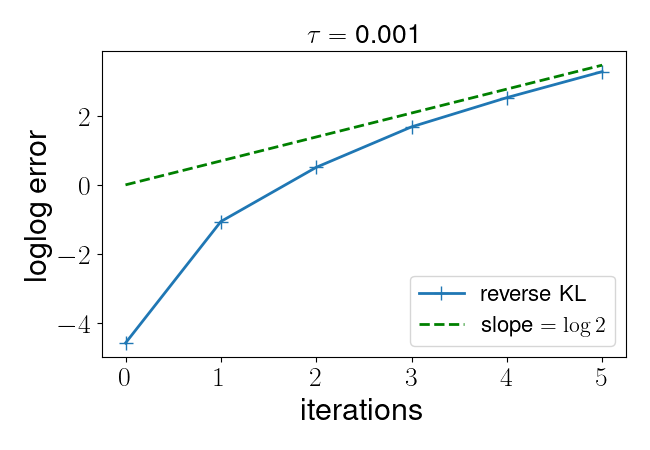} }
  \subfigure[The policy error in the process of training using Hellinger divergence. \label{fig:model1_Hellinger}]{
    \includegraphics[width=0.3\textwidth]{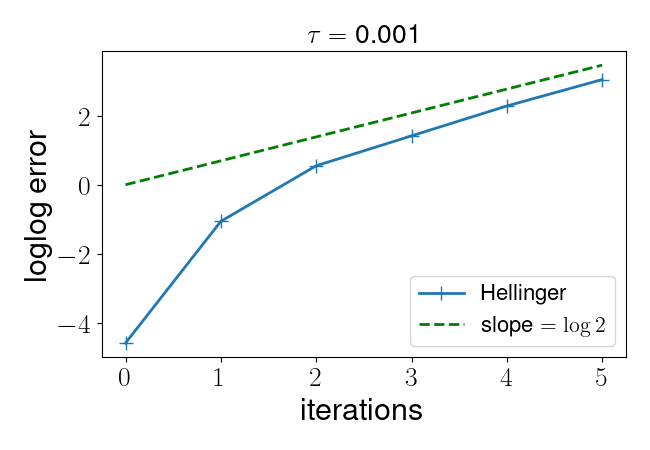} }
  \hspace{0.7em}
  \subfigure[The policy error in the process of training using $\alpha$-divergence with $\alpha=-3$. \label{fig:model1_alpha}]{
    \includegraphics[width=0.3\textwidth]{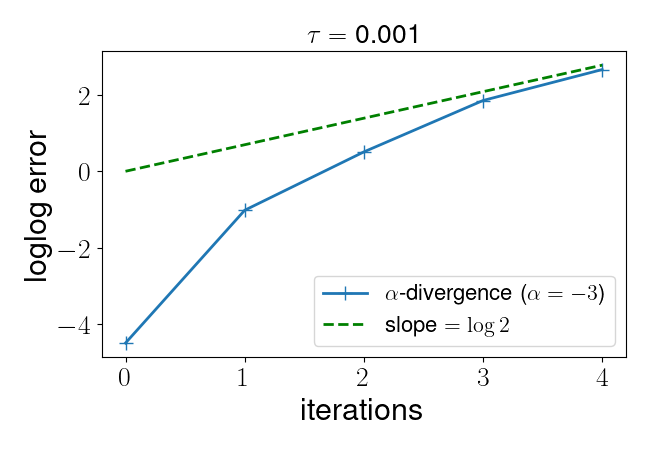} }
  \caption{Figures for the synthetic medium scale MDP. (a): Relative change of the policy
    $\norm{\pi_{\text{new}}-\pi}_F / \norm{\pi}_F$ during training of \Cref{alg:quasi} compared
    with PMD and GPMD in \cite{zhan2021policy}, with the logarithmic scale used for both axes.
    Notice that \Cref{alg:quasi} converges in 6-7 iterations to $10^{-12}$ in all cases
    while PMD and GPMD take more than $10^4$ iterations. Here the quadratic regularization is used for PMD and GPMD. (b) - (e): Blue: The convergence of
    $\log\left|\log\norm{\pi-\pi^*}_F\right|$ in the training process with the KL divergence, the
    reverse KL divergence, the Hellinger divergence, and the $\alpha$-divergence with $\alpha=-3$,
    respectively. Green: A line through the origin with slope $\log2$. Comparison of the convergence plots with the green reference lines shows a clear quadratic convergence for \Cref{alg:quasi}.}
  \label{fig:toy1quad}
\end{figure}

In the numerical experiment, we implement \Cref{alg:quasi} with the KL divergence, the
reverse KL divergence, the Hellinger divergence, and the $\alpha$-divergence with $\alpha=-3$. We adopt the uniform prior $\mu_s^a=1/|A|$ in order to make a fair comparison with the policy mirror descent (PMD) and the general policy mirror descent (GPMD) method in \cite{zhan2021policy}. We set the initial policy as the uniform policy, the convergence threshold as $\epsilon_{\text{tol}} = 10^{-12}$, and the learning rate $\eta$ as $1$. \Cref{fig:toy1} demonstrates that, for these four tests, the approximate Newton algorithm converges in $7$, $7$, $7$, and $6$ iterations, respectively. In comparison, we apply PMD and GPMD to the same MDP with the same stopping criterion. As also shown in
\Cref{fig:toy1}, many more iterations are needed for GPMD and PMD to reach the same precision:
GPMD converges in $14822$ iterations, and PMD does not reach the desired precision after
$3\times10^5$ iterations. For the implementation of GPMD and PMD, a quadratic regularization is used and we have already tuned the hyperparameters to optimize their performance. The number of iterations needed for GPMD and PMD to converge accords with the numerical results provided in \cite{zhan2021policy}.

\begin{sloppypar}
In order to verify the quadratic convergence proved in \cref{sec:analysis}, we draw the plots of 
$\log\left|\log\|\pi-\pi^*\|\right|$ in \Cref{fig:model1_KL}, \Cref{fig:model1_rKL}, \Cref{fig:model1_Hellinger} and \Cref{fig:model1_alpha}, where $\pi^*$ is the final policy and the norm used is the Frobenius norm. A green reference line with slope $\log2$ through the origin is plotted for comparison. If the error converges exactly at a quadratic rate, the plot of $\log\left|\log\|\pi-\pi^*\|\right|$ shall be parallel to the reference line. The convergence curves approach the reference lines at the end (and are even steeper than the reference lines in the beginning), demonstrating clearly a quadratic convergence for all forms of regularization used here. 
\end{sloppypar}

\subsection{Experiment II}

Next, we apply the approximate Newton methods derived in \cref{sec:Newton} to an MDP model
constructed from the search logs of an online shopping store, with two different ranking strategies.
Each issued query is represented as a state in the MDP. In response to a query, the search can be
done by choosing one of the two ranking strategies (actions) to return a ranked list of products
shown to the customer. Based on the shown products, the customer can refine or update the query,
thus entering a new state. The reward at each state-action pair is a weighted sum of the clicks and
purchases resulting from the action. Based on the data collected from two separate 5-week periods
for both ranking strategies, we construct an MDP with 135k states and a very sparse transition
tensor $P$ with only $0.01\%$ nonzero entries. The discount rate $\gamma$ is set as $0.99$ and the
regularization coefficient is $\tau=0.001$. In the implementation, we use the uniform prior $\mu_s^a=1/|A|$. 

When calculating $v_\pi$ by $v_\pi = (I-\gamma P_\pi)^{-1}(r_\pi-\tau h_\pi)$, we apply the
iterative solver Bi-CGSTAB \cite{van1992bi}, a widely used numerical method with high efficiency
and robustness for solving large sparse non-symmetric systems of linear equations
\cite{saad2003iterative,de1998comparison}, in order to leverage the sparsity of the transition
tensor.

\begin{figure}[ht]
  \centering
  \subfigure[Relative change of the policy $\norm{\pi_{\text{new}}-\pi}_F / \norm{\pi}_F$ in the training process. \label{fig:real}]{
    \includegraphics[width=0.3\textwidth]{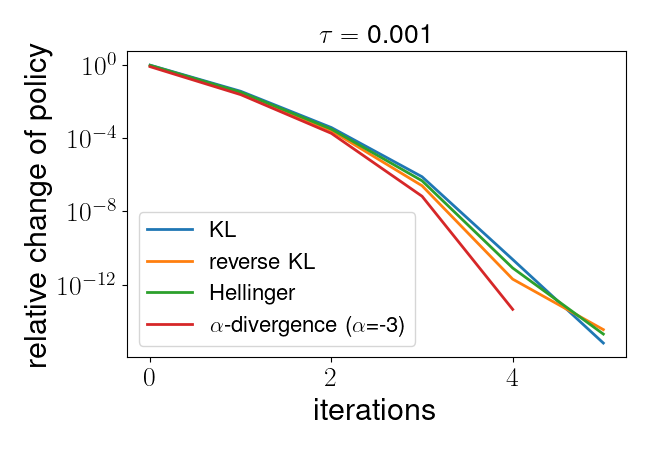} }
  \hspace{0.7em}
  \subfigure[The policy error in the process of training using KL-divergence. \label{fig:modelreal_KL}]{
    \includegraphics[width=0.3\textwidth]{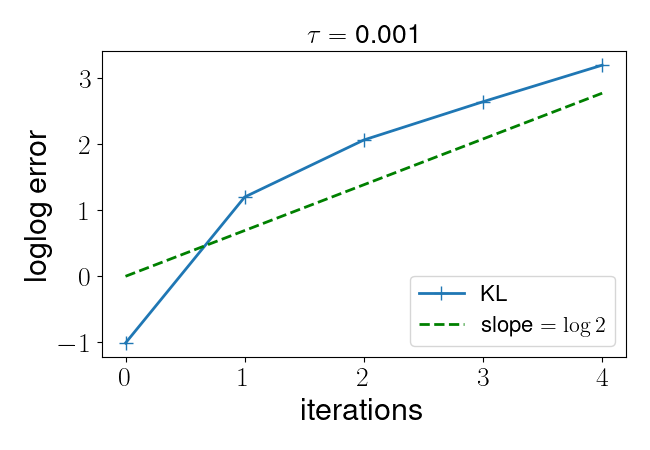} }
  \subfigure[The policy error in the process of training using reverse KL-divergence. \label{fig:modelreal_rKL}]{
    \includegraphics[width=0.3\textwidth]{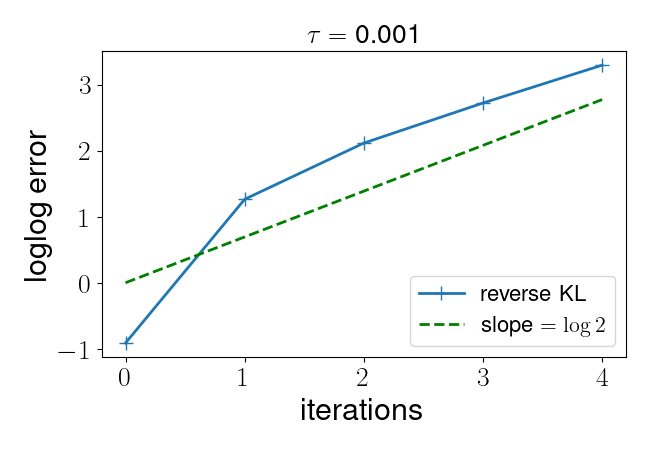} }
  \hspace{0.7em}
  \subfigure[The policy error in the process of training using Hellinger divergence. \label{fig:modelreal_Hellinger}]{
    \includegraphics[width=0.3\textwidth]{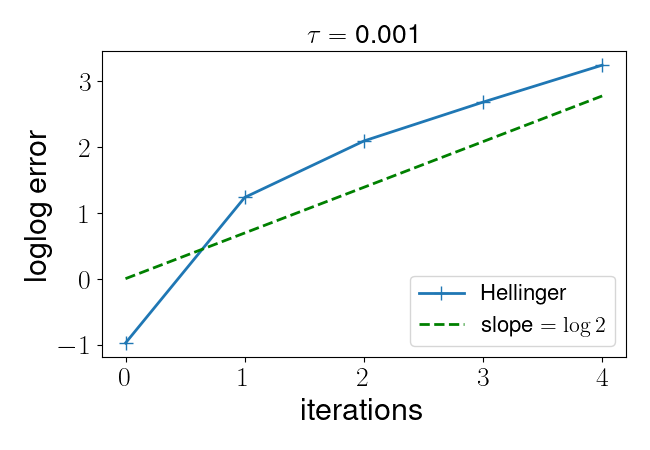} }
  \hspace{0.7em}
  \subfigure[The policy error in the process of training using $\alpha$-divergence with $\alpha=-3$. \label{fig:modelreal_alpha}]{
    \includegraphics[width=0.3\textwidth]{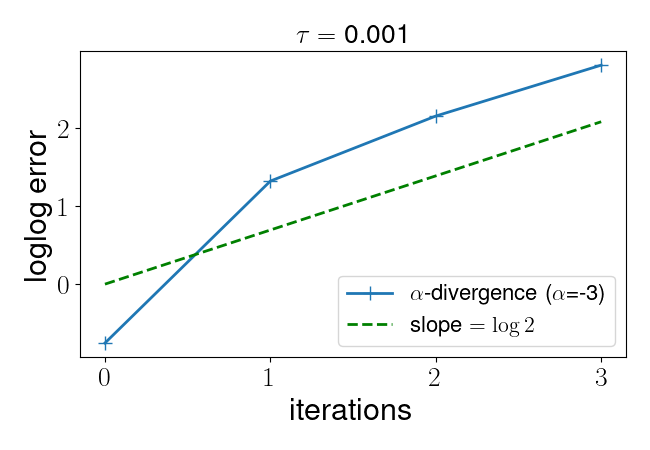} }
  \caption{Figures for the industrial-size MDP. (a): Relative change of the policy $\norm{\pi_{\text{new}}-\pi}_F / \norm{\pi}_F$ in the training process of \Cref{alg:quasi}. A logarithmic scale is used for the vertical axis. (b) - (e): Blue: The convergence of $\log\left|\log\norm{\pi-\pi^*}_F\right|$ in the training process with the KL divergence, the reverse KL divergence, the Hellinger divergence and the $\alpha$-divergence with $\alpha=-3$, respectively. Green: A line through the origin with slope $\log2$. }
  \label{fig:realquad}
\end{figure}

In the numerical experiment, we implement \Cref{alg:quasi} with the KL divergence, the reverse KL divergence, the Hellinger divergence, and the $\alpha$-divergence with $\alpha=-3$. We set the initial policy as the uniform policy, the convergence threshold as $\epsilon_{\text{tol}} = 10^{-12}$, and the learning rate $\eta$ as $1$. All the tests end up with fast convergence as shown in \Cref{fig:real}, where logarithmic scale is used for the vertical axis. More specifically, the approximate Newton algorithm using the KL divergence, the reverse KL divergence, the Hellinger divergence, and the $\alpha$-divergence with $\alpha=-3$ converge in $6, 6, 6, 5$ iterations, respectively. It is worth noticing that even though the size of the state space $S$ here is some magnitudes larger than the examples in \cref{sec:exptoy}, the number of approximate Newton iterations used is about the same. The comparison with GPMD and PMD is not given for this example since they are intractable to implement due to the high computational cost caused by the large MDP model. 

In \Cref{table:BiCGSTAB}, we report the number of BiCGSTAB steps used in the algorithm. In each
approximate Newton iteration, less than $20$ BiCGSTAB steps are used in order to find $v_\pi$. For all four regularizers used here, altogether only about $100$ BiCGSTAB steps are needed in the whole training process, thanks to the fast convergence of the approximate Newton method. As a comparison, the regularized value iteration (a special case for the method in \cite{geist2019theory}) typically needs thousands of matrix-vector multiplication with the MDP transition matrix, since its convergence rate is $O(\gamma^T)$. 

\begin{table}[ht]
\centering
\begin{tabular}{ cccc c  }
 \hline\hline
 Regularizer &\quad KL\quad&\quad reverse-KL\quad&\quad Hellinger\quad&$\alpha$-divergence ($\alpha=-3$)\\
 \hline\hline
 Approx-Newton Iterations   & \quad $6$\quad   & \quad $6$\quad & \quad $6$\quad & \quad $5$\quad \\
 \hline
 Total Bi-CGSTAB steps     & \quad $110$\quad & \quad $109$\quad & \quad $110$\quad & \quad $83$\quad \\
 \hline
 Average Bi-CGSTAB steps   &\quad $18.3$\quad & \quad $18.2$\quad & \quad $18.3$\quad & \quad $16.6$\quad \\
 \hline
 \vspace{1pt}
\end{tabular}
  \caption{Number of approximate Newton iterations and BiCGSTAB steps used in the training process. } \label{table:BiCGSTAB}
\end{table}

As in the previous numerical example, in \Cref{fig:modelreal_KL}, \Cref{fig:modelreal_rKL}, \Cref{fig:modelreal_Hellinger} and \Cref{fig:modelreal_alpha} we verify the quadratic convergence by comparing the plot of $\log\left|\log\|\pi-\pi^*\|\right|$ with a green reference line through the origin with slope $\log2$. As the convergence curves are approximately parallel to the reference lines, this verifies that the proposed algorithm converges quadratically with all the regularizations in this example as well.

\subsection{Experiment III}

In this section, we are concerned with an MDP with relatively large action space and state space at the same time. We consider the state space and action space with size $|S|=10000$ and $|A|=300$ with $(S, A)=(\{0,1,\ldots, |S|-1\}, \{0, 1, \ldots, |A|-1\})$. Here the transition tensor is defined as $P_{tt'}^a=1$ when $t'=(t+a) \bmod |S|, t\not=|S|-1 $ or $t=t'=|S|-1$, and $P_{tt'}^a=0$ otherwise. The reward is set as $r_s^a=1-\gamma$ if $s=|S|-1$ and $r_s^a=0$ otherwise, where $\gamma=0.99$. 

\begin{figure}[ht]
  \centering
  \subfigure[Relative change of the policy $\norm{\pi_{\text{new}}-\pi}_F / \norm{\pi}_F$ in the training process. \label{fig:large}]{
    \includegraphics[width=0.3\textwidth]{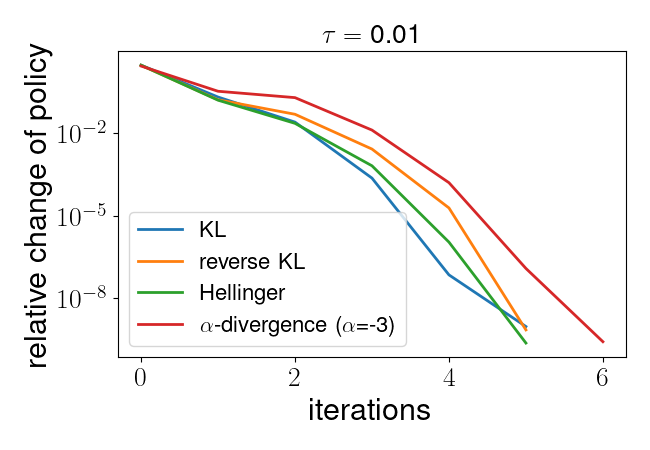} }
  \hspace{0.7em}
  \subfigure[The policy error in the process of training using KL-divergence. \label{fig:modellarge_KL}]{
    \includegraphics[width=0.3\textwidth]{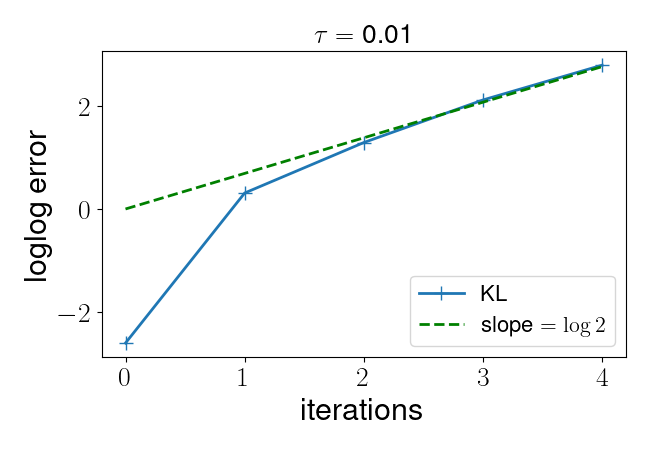} }
  \subfigure[The policy error in the process of training using reverse KL-divergence. \label{fig:modellarge_rKL}]{
    \includegraphics[width=0.3\textwidth]{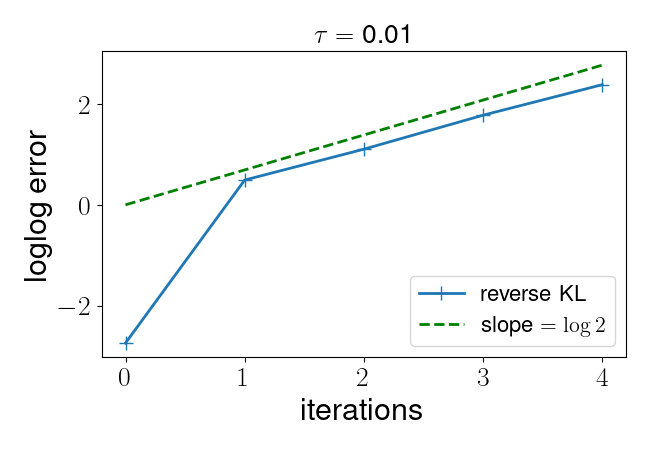} }
  \hspace{0.7em}
  \subfigure[The policy error in the process of training using Hellinger divergence. \label{fig:modellarge_Hellinger}]{
    \includegraphics[width=0.3\textwidth]{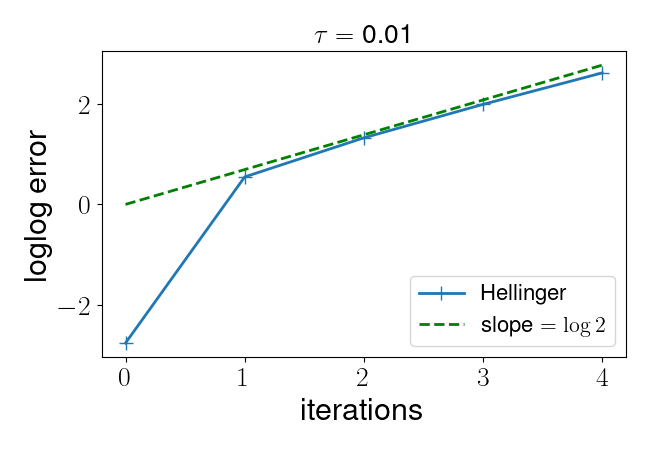} }
  \hspace{0.7em}
  \subfigure[The policy error in the process of training using $\alpha$-divergence with $\alpha=-3$. \label{fig:modellarge_alpha}]{
    \includegraphics[width=0.3\textwidth]{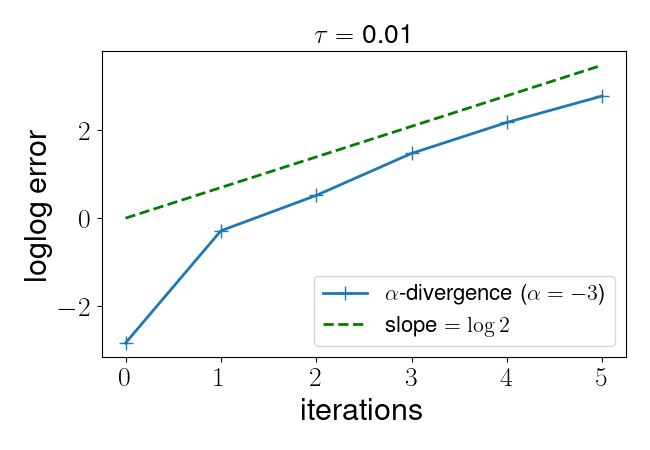} }
  \caption{Figures for an MDP with both the state space and action space relatively large. (a): Relative change of the policy $\norm{\pi_{\text{new}}-\pi}_F / \norm{\pi}_F$ in the training process of \Cref{alg:quasi}. A logarithmic scale is used for the vertical axis. (b) - (e): Blue: The convergence of $\log\left|\log\norm{\pi-\pi^*}_F\right|$ in the training process with the KL divergence, the reverse KL divergence, the Hellinger divergence and the $\alpha$-divergence with $\alpha=-3$, respectively. Green: A line through the origin with slope $\log2$. }
  \label{fig:largequad}
\end{figure}

Similar to the previous tests, we apply the approximate Newton algorithm with the KL divergence, the reverse KL divergence, the Hellinger divergence, and the $\alpha$-divergence with $\alpha=-3$ and the uniform prior $\mu_s^a=1/|A|$. For this experiment, we use $\tau=0.01$ and $\epsilon_{\text{tol}}=10^{-9}$. Similar to the previous examples, in all $4$ tests the algorithm converges with single-digit approximate Newton iterations, as shown in \Cref{fig:large}. The quadratic convergence can be verified in the plots of $\log\left|\log\|\pi-\pi^*\|\right|$ displayed in \Cref{fig:modellarge_KL}, \Cref{fig:modellarge_rKL}, \Cref{fig:modellarge_Hellinger} and \Cref{fig:modellarge_alpha}. Due to the size and sparsity of the transition tensor, we also adopt Bi-CGSTAB for calculating $v_\pi$, and the number of Bi-CGSTAB iterations used is reported in \Cref{table:BiCGSTABlarge}. Around $400$ Bi-CGSTAB steps are used, which also involves fewer matrix-vector multiplication with the transition matrix compared to traditional value iteration methods. 

\begin{table}[ht]
\centering
\begin{tabular}{ cccc c  }
 \hline\hline
 Regularizer &\quad KL\quad&\quad reverse-KL\quad&\quad Hellinger\quad&$\alpha$-divergence ($\alpha=-3$)\\
 \hline\hline
 Approx-Newton Iterations   & \quad $6$\quad   & \quad $6$\quad & \quad $6$\quad & \quad $7$\quad \\
 \hline
 Total Bi-CGSTAB steps     & \quad $370$\quad & \quad $379$\quad & \quad $492$\quad & \quad $452$\quad \\
 \hline
 Average Bi-CGSTAB steps   &\quad $61.7$\quad & \quad $63.2$\quad & \quad $82.0$\quad & \quad $64.6$\quad \\
 \hline
 \vspace{1pt}
\end{tabular}
  \caption{Number of approximate Newton iterations and BiCGSTAB steps used in the training process. } \label{table:BiCGSTABlarge}
\end{table}
\section{Proofs}\label{sec:proofs}

\subsection{Proof of \Cref{thm: approxnewton}}
\label{sec:approxnewton}
\begin{proof}
  \textbf{Step 1: expand $E(\pi)$ and prove the first-order condition \eqref{eq:grad}.}
  For any $\epsilon\in \R^{|S|\times|A|}$, introduce
  $r_\epsilon\in \R^{|S|}$ and $Z_\epsilon \in \R^{|S|\times|S|}$ such that
  \begin{equation}\label{eq:rZexpansion}
  (r_\epsilon)_s :=\sum_a\epsilon_s^ar_{s}^a, \quad
  (Z_\epsilon)_{st}:=\sum_a\epsilon_s^a(\delta_{st}-\gamma P_{st}^a),
  \end{equation}
  where $\delta_{st} = 1$ if $s=t$ and $\delta_{st} = 0$ otherwise. Then $r_\epsilon$ and $Z_\epsilon$ are linear with respect to $\epsilon$, which is helpful when expressing the first-order conditions and simplifying the expansion of $E(\pi)$. 
  
  Now we proceed to prove that for any $\epsilon$ with $\sum_a\epsilon_s^a = 0$ and $|\epsilon_s^a|<\pi_s^a$, at $\pi=\pi^*$
  \begin{equation}\label{eq:grad}
    r_\epsilon-\tau Dh_\pi\epsilon-Z_\epsilon Z_\pi^{-1}(r_\pi-\tau h_\pi) = 0,
  \end{equation}
  where $Dh_\pi \in \R^{|S|\times |S||A|}$ is the gradient matrix of $h_\pi$ with respect to
  $\pi$.
  
  Since $\pi$ is a policy, $\sum_a\pi_s^a = 1$ for any $s$. Thus
  \begin{equation}\label{eq:linear0}
    (Z_\pi)_{st} = \delta_{st} - \gamma\sum_a\pi_s^aP_{st}^a = \sum_a\pi_s^a(\delta_{st}-\gamma P_{st}^a).
  \end{equation}
  Now consider a policy $\pi+\epsilon$, i.e., $\sum_a\epsilon_s^a = 0$ and
  $\pi_s^a+\epsilon_s^a\geq0$, then thanks to \eqref{eq:linear0} one can obtain:
  \begin{equation}\label{eq:linear}
    Z_{\pi+\epsilon} = Z_\pi+Z_\epsilon, \quad r_{\pi+\epsilon} = r_\pi+r_\epsilon,
  \end{equation}
  where $Z_\epsilon$ and $r_\epsilon$ are defined in \eqref{eq:rZexpansion}, i.e.,
  $(Z_\epsilon)_{st}=\sum_a\epsilon_s^a(\delta_{st}-\gamma P_{st}^a)$. $(r_\epsilon)_s =\sum_a\epsilon_s^ar_{s}^a$.
  Leveraging the linearity \eqref{eq:linear}, we obtain the expansion:
  \begin{equation}\label{eq:expansion}
    \begin{aligned}
      &\quad E(\pi+\epsilon) = \rho^\top Z_{\pi+\epsilon}^{-1}(r_{\pi+\epsilon}-\tau
      h_{\pi+\epsilon}) =\rho^\top (Z_{\pi}+Z_{\epsilon})^{-1}(r_{\pi}+r_{\epsilon}-\tau
      h_{\pi+\epsilon})\\ &=\rho^\top Z_{\pi}^{-1}
      (I-Z_{\epsilon}Z_\pi^{-1}+Z_{\epsilon}Z_\pi^{-1}Z_{\epsilon}Z_\pi^{-1})(r_{\pi}+r_{\epsilon}-\tau
      h_{\pi}-\tau Dh_\pi\epsilon-\frac{1}{2}\epsilon^\top \tau D^2h_\pi\epsilon) + O(\|\epsilon\|^3)\\ 
      &= E(\pi) + w_\pi^\top\left[-Z_{\epsilon}Z_\pi^{-1}(r_\pi-\tau h_\pi)+(r_\epsilon-\tau Dh_\pi\epsilon)\right]+w_\pi^\top(-\frac{1}{2}\epsilon^\top \tau
        D^2h_\pi\epsilon)\\ 
      &\quad+w_\pi^\top\left[-Z_{\epsilon}Z_\pi^{-1}(r_\epsilon-\tau
        Dh_\pi\epsilon)+Z_{\epsilon}Z_\pi^{-1}Z_{\epsilon}Z_\pi^{-1}(r_\pi-\tau
        h_\pi)\right] + O(\|\epsilon\|^3),
    \end{aligned}
  \end{equation}
  where $Dh_\pi$ is a second-order tensor that maps from $S\times A$ to $S$, and $D^2h_\pi$ is a
  third-order tensor that maps from $(S\times A)^{\otimes2}$ to $S$. With this expansion, one can see
  that
  \[
    \frac{\partial E}{\partial \pi_s^a} = (r_s^a - \tau (\log\pi_s^a +1) - \left[(I-\gamma P^a)v_\pi\right]_s + c_s)(w_\pi)_s,
  \]
  where $c_s$ is a multiplier that only depends on $s$. 
  Then at $\pi = \pi^*$,
  \[
  \frac{\partial E}{\partial \pi_s^a} = (r_s^a - \tau (\log\pi_s^a + 1) - \left[(I-\gamma P^a)v_\pi\right]_s + c_s)(w_\pi)_s = 0.
  \]
  Since $w_\pi = (I-\gamma P_\pi^\top)^{-1}\rho = \rho + \sum_{i=1}^\infty \gamma^i(P_\pi^\top)^ie$ and
  all elements of $\rho$ are positive, we also know that all elements of $w_\pi$ are positive. Thus at
  $\pi = \pi^*$,
  \[
  r_s^a - \tau (\log\pi_s^a + 1) - \left[(I-\gamma P^a)v_\pi\right]_s + c_s = 0. 
  \]
  Multiplying the left hand side with $\epsilon_s^a$ and taking the sum over $a$, we obtain: 
  \[
  \left(r_\epsilon-\tau Dh_\pi\epsilon-Z_\epsilon v_\pi\right)_s + c_s\sum_a\epsilon_s^a = 0, \quad \forall s, \quad\forall \epsilon.
  \]
  Since $\sum_a\epsilon_s^a = 0$ for any $s$ and $v_\pi = Z_\pi^{-1}(r_\pi-\tau h_\pi)$, we have 
  \[
  r_\epsilon-\tau Dh_\pi\epsilon-Z_\epsilon Z_\pi^{-1}(r_\pi-\tau h_\pi) = 0, \quad\forall \epsilon,
  \]
  at $\pi = \pi^*$, which proves \eqref{eq:grad}.
  
  \textbf{Step 2: Derive the decomposition \eqref{eq:decomp} with the obtained expansion and first-order condition.}
  With \eqref{eq:grad}, one can approximate the second-order term in \eqref{eq:expansion} for $\pi$ near $\pi^*$:
  \[
    \begin{aligned}
      &w_\pi^\top(-\frac{1}{2}\epsilon^\top \tau D^2h_\pi\epsilon)+w_\pi^\top\left[-Z_{\epsilon}Z_\pi^{-1}(r_\epsilon-\tau Dh_\pi\epsilon)+Z_{\epsilon}Z_\pi^{-1}Z_{\epsilon}Z_\pi^{-1}(r_\pi-\tau h_\pi)\right]\\
      &=\frac{1}{2}\epsilon^\top\Lambda(\pi)\epsilon-w_\pi^\top Z_{\epsilon}Z_\pi^{-1}\left(r_\epsilon-\tau Dh_\pi\epsilon-Z_{\epsilon}Z_\pi^{-1}(r_\pi-\tau h_\pi)\right)\\
      &\approx \frac{1}{2}\epsilon^\top\Lambda(\pi)\epsilon.
    \end{aligned}
  \]
  By \eqref{eq:grad} and that $h$ is twice continuously differentiable, the approximate Hessian $\Lambda$ converge to the true Hessian as $\pi$ converges to $\pi^*$, and their difference $\Lambda(\pi) - D^2E(\pi) = O(\|\pi-\pi^*\|)$. Hence, the second-order derivatives of $E(\pi)$ can be approximated by
  \begin{equation}
    \frac{\partial^2 E}{\partial \pi_s^a\partial \pi_t^b}\approx \Lambda_{(sa),(tb)} = -\tau \delta_{\{(sa),(tb)\}}\frac{(w_\pi)_s}{\pi_s^a},
  \end{equation}
  from which we have shown that $\Lambda$ is diagonal.

  \textbf{Step 3: Derive the approximate Newton flow and the policy update scheme with the obtained decomposition.}
  Using this approximate second-order derivative as a preconditioner, $w_\pi$ is canceled out in the
  policy gradient algorithm, which yields the gradient flow:
  \[
  \frac{\mathrm{d}\pi_s^a}{\mathrm{d} t} = \pi_s^a(r_s^a-\tau(\log\pi_s^a+1)-[(I-\gamma P^a)v_\pi]_s + c_s)/\tau.
  \]
  Adopting the parameterization $\pi_s^a = \exp(\theta_s^a)$, we have
  \begin{equation}\label{eq:paramgrad}
    \frac{\mathrm{d}\theta_s^a}{\mathrm{d} t} = (r_s^a-\tau(\theta_s^a+1)-[(I-\gamma P^a)v_\pi]_s + c_s)/\tau.
  \end{equation}
  With a learning rate $\eta$, this becomes
  \begin{equation}
    \theta_s^a \leftarrow \eta(r_s^a-\tau-[(I-\gamma P^a)v_\pi]_s + c_s)/\tau + (1-\eta)\theta_s^a,
  \end{equation}
  which corresponds to
  \begin{equation}
    \pi_s^a \leftarrow (\pi_s^a)^{1-\eta}\exp(\eta(r_s^a-\tau-[(I-\gamma P^a)v_\pi]_s + c_s)/\tau),
  \end{equation}
  and $c_s$ is determined by the condition that $\sum_a\pi_s^a=1$. Equivalently, we have
  \[
  \pi_s^a \leftarrow \frac{(\pi_s^a)^{1-\eta}\exp(\eta(r_s^a-\tau-[(I-\gamma P^a)v_\pi]_s)/\tau)}{\sum_a (\pi_s^a)^{1-\eta}\exp(\eta(r_s^a-\tau-[(I-\gamma P^a)v_\pi]_s)/\tau)} = \frac{(\pi_s^a)^{1-\eta}\exp(\eta(r_s^a+(\gamma P^av_\pi)_s)/\tau)}{\sum_a(\pi_s^a)^{1-\eta}\exp(\eta(r_s^a+(\gamma P^av_\pi)_s)/\tau)},
  \]
  where we cancel out the factors independent of $a$ and obtain \eqref{eq:NPG}. This finishes the proof.
\end{proof}

\subsection{Proof of \Cref{thm: otheretp}}
\label{sec:otheretp}
\begin{proof}
  Similar with \eqref{eq:grad}, we first prove that for any $\epsilon$ with $\sum_a\epsilon_s^a = 0$ and $|\epsilon_s^a|<\pi_s^a$, at $\pi=\pi^*$
  \begin{equation}\label{eq:general1stmain}
    r_\epsilon-\tau Dh_\pi\epsilon-Z_\epsilon Z_\pi^{-1}(r_\pi-\tau h_\pi) = 0,
  \end{equation}
  
  Similar to the proof of \Cref{thm: approxnewton}, by direct calculations one can get:
  \begin{equation}
    \frac{\partial E}{\partial \pi_s^a} = (r_s^a - \tau \phi'(\pi_s^a/\mu_s^a) - \left[(I-\gamma P^a)v_\pi\right]_s + c_s)(w_\pi)_s,
  \end{equation}
  where $c_s$ is a multiplier that only depends on $s$. Since all elements of $w_\pi$ are positive, at $\pi = \pi^*$,
  \begin{equation}\label{eq:general1st}
    (r_s^a - \tau \phi'(\pi_s^a/\mu_s^a) - \left[(I-\gamma P^a)v_\pi\right]_s + c_s) = 0.
  \end{equation}
  By multiplying \eqref{eq:general1st} with $\epsilon_s^a$ and summing over $a$, one can obtain:
  \[
  r_\epsilon-\tau Dh_\pi\epsilon-Z_\epsilon Z_\pi^{-1}(r_\pi-\tau h_\pi) = 0, \quad\forall \epsilon,
  \]
  at $\pi = \pi^*$, which proves \eqref{eq:general1stmain}. 
  Since the only difference between the functional $E(\pi)$ defined here and the $E(\pi)$ in \Cref{thm: approxnewton} lies in the regularizer $h$, one can still obtain the expansion:
  \[
    \begin{aligned}
    &E(\pi+\epsilon) - E(\pi) - w_\pi^\top\left[-Z_{\epsilon}Z_\pi^{-1}(r_\pi-\tau h_\pi)+(r_\epsilon-\tau Dh_\pi\epsilon)\right]\\
    &= w_\pi^\top(-\frac{1}{2}\epsilon^\top \tau D^2h_\pi\epsilon) -w_\pi^\top Z_{\epsilon}Z_\pi^{-1}\left(r_\epsilon-\tau Dh_\pi\epsilon-Z_{\epsilon}Z_\pi^{-1}(r_\pi-\tau h_\pi)\right) + O(\|\epsilon\|^3)\\
    &= \frac{1}{2}\epsilon^\top\Lambda(\pi)\epsilon + O(\|\epsilon\|^2\|\pi-\pi^*\|) + O(\|\epsilon\|^3).
    \end{aligned}
  \]
  Hence we have $D^2E(\pi)-\Lambda(\pi) = O(\|\pi-\pi^*\|)$. Using this expansion, one can derive an approximation for the second-order derivatives:
  \[
    \frac{\partial^2 E}{\partial \pi_s^a\partial \pi_t^b} \approx \Lambda_{(sa),(tb)} = -\tau\delta_{\{(sa),(tb)\}}\frac{(w_\pi)_s\phi''(\pi_s^a/\mu_s^a)}{\mu_s^a},
  \]
  which proves \eqref{eq:generalapproxhess} and shows that $\Lambda$ is diagonal. The approximate Newton flow thus becomes
  \[
    \frac{\mathrm{d}\pi_s^a}{\mathrm{d} t} = \mu_s^a(\phi''(\pi_s^a/\mu_s^a))^{-1}(r_s^a- \tau \phi'(\pi_s^a/\mu_s^a)-[(I-\gamma P^a)v_\pi]_s+c_s)/\tau,
  \]
  which proves \eqref{eq:qNflowgen}, or equivalently,
  \eq{
    \frac{\mathrm{d}(\phi'(\pi_s^a/\mu_s^a))}{\mathrm{d} t} = (r_s^a- \tau \phi'(\pi_s^a/\mu_s^a)-[(I-\gamma P^a)v_\pi]_s+c_s)/\tau.
  }
  Let $\theta_s^a = \phi'(\pi_s^a/\mu_s^a)$, then
  \[
  \frac{\mathrm{d}\theta_s^a}{\mathrm{d} t} = (r_s^a-\tau\theta_s^a-[(I-\gamma P^a)v_\pi]_s+c_s)/\tau,
  \]
  With a learning rate $\eta$, this becomes
  \[
    \theta_s^a \leftarrow \eta(r_s^a-[(I-\gamma P^a)v_\pi]_s + c_s)/\tau + (1-\eta)\theta_s^a,
  \]
  which proves \eqref{eq:quasiNTupdate}. 

\end{proof}

\subsection{Proof of \Cref{lemma:multipliergen}}
\label{sec:mult}

\begin{proof}
  Let
  \[
  g(x) = \mu_1\psi(x+x_1) + \cdots + \mu_k\psi(x+x_k).
  \]
  Since $\psi:(L, +\infty)\rightarrow(0, +\infty)$ is decreasing, $g(x)$ is positive and decreasing on $(L-\underset{1\leq i\leq k}{\min}x_i, +\infty)$. When $x\rightarrow-\underset{1\leq i\leq k}{\min}x_i$
  from the right, $g(x)\rightarrow+\infty$ since at least one of the terms go to $+\infty$. If
  $\underset{1\leq i\leq k}{\min}\left\{\psi^{-1}(\frac{1}{k\mu_i}) - x_i\right\}\geq L-\underset{1\leq i\leq k}{\min}x_i$, when $x
  = \underset{1\leq i\leq k}{\min}\left\{\psi^{-1}(\frac{1}{k\mu_i}) - x_i\right\}$
  \[
  \begin{aligned}
  g(x) &= \sum_{i=1}^k\mu_i\psi\left(\underset{1\leq j\leq k}{\min}\left\{\psi^{-1}\left(\frac{1}{k\mu_j}\right) - x_j\right\}+x_i\right)\\
  &= \sum_{i=1}^k\mu_i\psi\left(\underset{1\leq j\leq k}{\min}\left\{\psi^{-1}\left(\frac{1}{k\mu_j}\right) - x_j\right\}+x_i-\psi^{-1}\left(\frac{1}{k\mu_i}\right)+\psi^{-1}\left(\frac{1}{k\mu_i}\right)\right)\\
  &\geq \sum_{i=1}^k\mu_i\psi\left(\psi^{-1}\left(\frac{1}{k\mu_i}\right)\right) =\sum_{i=1}^k \mu_i\times\frac{1}{k\mu_i} = k \times \frac{1}{k} = 1.
  \end{aligned}
  \]
  Since $\psi^{-1}\left(\frac{1}{k\mu_i}\right)\geq L$, we have $\underset{1\leq i\leq k}{\max}\left\{\psi^{-1}\left(\frac{1}{k\mu_i}\right) - x_i\right\}\geq \underset{1\leq i\leq k}{\max}\left\{L - x_i\right\} = L - \underset{1\leq i\leq k}{\min}x_i$. 
  Then when $x = \underset{1\leq i\leq k}{\max}\left\{\psi^{-1}\left(\frac{1}{k\mu_i}\right) - x_i\right\}$,
  \[
  g(x) = \sum_{i=1}^k\mu_i\psi\left(\underset{1\leq j\leq k}{\max}\left\{\psi^{-1}\left(\frac{1}{k\mu_j}\right) - x_j\right\}+x_i\right)\leq \sum_{i=1}^k\mu_i\psi\left(\psi^{-1}\left(\frac{1}{k\mu_i}\right)\right) = 1.
  \]
  By the continuity of $g$, there exists a solution $x$ to \eqref{eq:lemmagen} on
  \[
  \left[\max\left\{L-\underset{1\leq i\leq k}{\min}x_i, \underset{1\leq i\leq k}{\min}\left\{\psi^{-1}\left(\frac{1}{k\mu_i}\right) - x_i\right\}\right\}, \underset{1\leq i\leq k}{\max}\left\{\psi^{-1}\left(\frac{1}{k\mu_i}\right) - x_i\right\}\right],
  \]
  and the solution is unique by the strict monotonicity of $g$ on $(L-\underset{1\leq i\leq k}{\min}x_i,
  \infty)$. 
\end{proof}

\subsection{Proof of \Cref{prop:multiplier}}
\label{sec:multsolve}
\begin{proof}
By \Cref{lemma:multipliergen} there is a unique solution $\tilde{c}_s$ to the equation $\sum_a \mu_s^a \psi(\tilde{c}_s + x_a) = 1$, where $x_a$ is defined as in \eqref{eq:deftildecxa}. 
Now update the policy by 
\[
\pi_s^a\leftarrow \mu_s^a \psi(\tilde{c}_s + x_a) = \sum_a\mu_s^a\psi\left(-\frac{\eta}{\tau}c_s-(1-\eta)\phi'\left(\pi_s^a/\mu_s^a\right) -\frac{\eta}{\tau}(r_s^a-[(I-\gamma P^a)v_\pi]_s)\right)
\]
one ensures that $\pi_s^a\geq0$ and $\sum_a\pi_s^a=1$, and that the multiplier $c_s$ with this property is unique. 

\end{proof}

\subsection{Proof of \Cref{thm:flowconv}}
\label{sec:flowconvpf}
\begin{proof}
In \cref{sec:flowconv} we have proved that the approximate Newton flow:
\[
\frac{\mathrm{d}\pi_s^a}{\mathrm{d} t} = \mu_s^a(\phi''(\pi_s^a/\mu_s^a))^{-1}(r_s^a- \tau \phi'(\pi_s^a/\mu_s^a)-[(I-\gamma P^a)v_\pi]_s+c_s)/\tau
\]
converges globally, so it suffices to show that the limiting policy is the optimal policy. Denote the limiting policy by $\pi^\diamond$. Since $\mu_s^a>0$ and $(\phi''((\pi^\diamond)_s^a/\mu_s^a))^{-1}>0$, we have 
\begin{equation}\label{eq:flow1st}
    r_s^a- \tau \phi'((\pi^\diamond)_s^a/\mu_s^a)-[(I-\gamma P^a)v_{\pi^\diamond}]_s+c_s = 0,
\end{equation}
and $c_s$ is a multiplier which ensures $\sum_a(\pi^\diamond)_s^a = 1$. From the theory of regularized MDP (Cf. \cite{geist2019theory}), we know that the optimal policy $\pi^*$ is the unique solution to the Bellman maximal equation:
\begin{equation}\label{eq:regBell}
    v = \underset{\pi}{\max}~ r_\pi + \g P_\pi v - \tau h_\pi.
\end{equation}
Since $v_{\pi^\diamond} = (I-\g P_{\pi^\diamond})^{-1}(r_{\pi^\diamond}-\tau h_{\pi^\diamond})$, we have $v_{\pi^\diamond}-\g P_{\pi^\diamond}v_{\pi^\diamond} = r_{\pi^\diamond}-\tau h_{\pi^\diamond}$, or equivalently 
\[v_{\pi^\diamond} = r_{\pi^\diamond} + \g P_{\pi^\diamond}v_{\pi^\diamond}-\tau h_{\pi^\diamond}.\]
Thus it now suffices to show that $\pi^\diamond$ is the optimizer of the constrained maximization problem $\underset{\pi}{\max}~ r_\pi + \g P_\pi v_{\pi^\diamond} - \tau h_\pi$, or in the component form:
\begin{equation}\label{eq:optprob}
  \underset{\pi}{\max} \sum_a (r_s^a+\g (P^av_{\pi^\diamond})_s)\pi_s^a - \tau \sum_a\mu_s^a\phi(\pi_s^a/\mu_s^a).
\end{equation}
Since $\phi$ is convex and $\mu$ is positive, $\tau \sum_a\mu_s^a\phi(\pi_s^a/\mu_s^a)$ is also a convex function in $\pi_s$. By the theory of convex optimization (Cf. \cite{boyd2004convex}, chapter $5$), the Karush-Kuhn-Tucker (KKT) condition is sufficient for the optimality when the objective function is convex, and the KKT condition for the problem \eqref{eq:optprob} is 
\[
\begin{aligned}
r_s^a+\g (P^av_{\pi^\diamond})_s -\tau \phi'(\pi_s^a/\mu_s^a) + \lambda_s &=0,\\
\sum_a\pi_s^a&=1,\\
\pi_s^a&\geq0,
\end{aligned}
\]
where $\lambda_s$ is the Lagrange multiplier. Now let $\pi=\pi^{\diamond}$ and $\lambda_s = c_s-(v_{\pi^\diamond})_s$. From the first-order condition \eqref{eq:flow1st} one can directly observe that the KKT condition above is satisfied, which makes $\pi^\diamond$ the optimizer for \eqref{eq:optprob} and $v_{\pi^\diamond}$ the solution to the Bellman equation \eqref{eq:regBell}. Thus $v_{\pi^\diamond}$ and $\pi^\diamond$ are indeed the optimal value function and the optimal policy, which closes the proof. 
\end{proof}

\subsection{Proof of \Cref{thm:quadconv}}

\label{sec:quadconvpf}
\begin{proof}
The proof is divided into three steps. First, we present some results needed in proving the local convergence. We then demonstrate the local convergence of $\pi\sps{k}$ to $\pi^*$ using induction in the second step. Finally, we prove that the convergence rate is quadratic. 

\textbf{Step 1. Preparation.} 
From the scheme
\eq{\label{eq:replace}
\grad\Phi(\pi\sps{k+1})-\grad\Phi(\pi\sps{k}) = -\left(f(\pi\sps{k}) + \grad\Phi(\pi\sps{k}) -
  B^\top c(\pi\sps{k})\right),
}
one can obtain the inequality
\begin{equation}\label{eq:updateineq}
\begin{aligned}
&{\|f(\pi\sps{k+1})-f(\pi\sps{k})\|}\geq \frac{(f(\pi\sps{k})-f(\pi\sps{k+1}))^\top(\pi\sps{k+2}-\pi\sps{k+1})}{\|\pi\sps{k+2}-\pi\sps{k+1}\|}\\
&=\frac{-\left(f(\pi\sps{k+1})+\grad\Phi(\pi\sps{k+1})-B^\top c(\pi\sps{k})\right)^\top(\pi\sps{k+2}-\pi\sps{k+1})}{\norm{\pi\sps{k+2}-\pi\sps{k+1}}}\\
&=\frac{-\left(f(\pi\sps{k+1})+\grad\Phi(\pi\sps{k+1})-B^\top c(\pi\sps{k+1})\right)^\top(\pi\sps{k+2}-\pi\sps{k+1})}{\norm{\pi\sps{k+2}-\pi\sps{k+1}}}\\
&=\frac{\left(\grad\Phi(\pi\sps{k+2})-\grad\Phi(\pi\sps{k+1})\right)^\top(\pi\sps{k+2}-\pi\sps{k+1})}{\norm{\pi\sps{k+2}-\pi\sps{k+1}}},
\end{aligned}
\end{equation}
where we use the constraint $B\pi\sps{k+1} = B\pi\sps{k+2} = \mathbf{1}_{|S|}$. 
By a direct calculation of $\grad^2\Phi$ from the definition of $\Phi$, we can see that $\grad^2\Phi$ is diagonal and $\Phi$ is strongly convex since $\phi$ is strongly convex.
As a result,
there is some constant $\omega>0$ such that
\begin{equation}\label{eq:strong}
(\grad\Phi(\pi)-\grad\Phi(\tilde{\pi}))^\top(\pi-\tilde{\pi}) \geq \omega\norm{\pi-\tilde{\pi}}^2
\end{equation}
for any $\pi$ and $\tilde{\pi}$. Thus from \eqref{eq:updateineq} one can deduce that
\begin{equation}\label{eq:fpiineq}
\begin{aligned}
\norm{f(\pi\sps{k+1})-f(\pi\sps{k})}&\geq \frac{\left(\grad\Phi(\pi\sps{k+2})-\grad\Phi(\pi\sps{k+1})\right)^\top(\pi\sps{k+2}-\pi\sps{k+1})}{\norm{\pi\sps{k+2}-\pi\sps{k+1}}} \\
&\geq \frac{\omega\norm{\pi\sps{k+2}-\pi\sps{k+1}}^2}{\norm{\pi\sps{k+2}-\pi\sps{k+1}}}={\omega\norm{\pi\sps{k+2}-\pi\sps{k+1}}}.
\end{aligned}
\end{equation}
Let $K$ be a closed set contained in $\{\pi:B^\top\pi=\mathbf{1}_{|S|}, \pi_s^a>0\}$ such that $K$ contains a ball centered at $\pi^*$ with radius $\delta_0>0$, which is guaranteed to exist since $(\pi^*)_s^a>0$.  
Define the conjugate function of $\Phi$ as
\eq{
  \Phi^*(x) = \max_{\pi\in\Delta} \left[\sum_{sa}\pi_{sa}x_{sa}-\Phi(\pi)\right],
}
where $\Delta = \{\pi:B^\top \pi=\mathbf{1}_{|S|}, \pi_{sa}\geq0\}$. Since $\Phi$ is $\omega$-strongly convex and $\Delta$ is a closed convex set, it can be deduced from classical convex analysis results (see \cite{hiriart2004fundamentals} for example) that $\grad\Phi^*$ is $\frac{1}{\omega}$-Lipschitz continuous, and $\pi=\grad\Phi^*(\grad\Phi(\pi))$. Moreover, from the definition of $\Phi^*$ one can observe that $\Phi^*(x+B^\top c) = \Phi^*(x)+\mathbf{1}_{|S|}^\top c$, and thus $\grad\Phi^*(x+B^\top c) = \grad\Phi^*(x)$. Similar results concerning the conjugate functions have also been used in \cite{mensch2018differentiable} and \cite{geist2019theory}. Thanks to the properties of $\Phi^*$, we have the identity
\eq{\label{eq:conjinvk}
  \pi\sps{k+1}=\grad\Phi^*(\grad\Phi(\pi\sps{k+1}))=\grad\Phi^*(B^\top c(\pi\sps{k})-f(\pi\sps{k}))=\grad\Phi^*(-f(\pi\sps{k})),
}
where we have used the update scheme \eqref{eq:replace}. 
Moreover, by the result of \Cref{thm:flowconv} we have $\frac{\mathrm{d}(\grad\Phi(\pi))}{\mathrm{d} t}=0$ at $\pi=\pi^*$, so $f(\pi^*)+\grad\Phi(\pi^*) = B^\top c(\pi^*)$ and
\eq{\label{eq:conjinvstar}
  \pi^*=\grad\Phi^*(\grad\Phi(\pi^*))=\grad\Phi^*(B^\top c(\pi^*)-f(\pi^*))=\grad\Phi^*(-f(\pi^*)),
}
Since $\grad\Phi^*$ and $-f$ are continuous on $K$, it can be concluded from \eqref{eq:conjinvk} and \eqref{eq:conjinvstar} that there exists $\delta_1>0$ such that $\norm{\pi\sps{k+1}-\pi^*}<\frac{1}{16}\min\{\frac{\omega}{M}, \delta_0\}$ whenever $\norm{\pi\sps{k}-\pi^*}\leq\delta_1$, where $M=\sup_{\pi\in K}|\grad^2 f(\pi)|$. 

\textbf{Step 2. Prove the convergence by induction.} 
Now let $\delta=\min\{\frac{\omega}{16M}, \frac{\delta_0}{16}, \delta_1\}$. Assuming that $\norm{\pi\sps{0}-\pi^*}<\delta$, we proceed to prove that $\norm{\pi\sps{k}-\pi^*}\leq \frac{1}{2}\min\{\frac{\omega}{M}, \delta_0\}$ for any $k$ by induction. To this end, we first strengthen the induction hypothesis to
\eq{\label{eq:induce}
  \begin{aligned}
  \norm{\pi\sps{k}-\pi^*}&\leq (\frac{1}{2}-\frac{1}{2^{k+2}})\min\{\frac{\omega}{M}, \delta_0\}, \quad k=0, 1, \ldots, n,\\
  \norm{\pi\sps{k+1}-\pi\sps{k}}&\leq (\frac{1}{2}-\frac{1}{2^{k+2}})\norm{\pi\sps{k}-\pi\sps{k-1}}, \quad k=1, 2, \ldots, n.\\
  \end{aligned}
}
We first prove \eqref{eq:induce} for $n=1$. Note that
\eq{\label{eq:pi0}
\norm{\pi\sps{0}-\pi^*}\leq\delta\leq (\frac{1}{2}-\frac{1}{2^{0+2}})\min\{\frac{\omega}{M}, \delta_0\},
}
by the definition of $\delta$, and that
\eq{\label{eq:pi1}
\norm{\pi\sps{1}-\pi^*}\leq\frac{1}{16}\min\{\frac{\omega}{M}, \delta_0\}\leq (\frac{1}{2}-\frac{1}{2^{1+2}})\min\{\frac{\omega}{M}, \delta_0\},
}
by the definition of $\delta_1$ and the fact that $\norm{\pi\sps{0}-\pi^*}\leq\delta_1$. Then
\eq{
\norm{\pi\sps{1}-\pi\sps{0}}\leq\norm{\pi\sps{1}-\pi^*}+\norm{\pi\sps{0}-\pi^*}\leq\frac{1}{8}\min\{\frac{\omega}{M}, \delta_0\}.
}
In addition, from \eqref{eq:pi0} and \eqref{eq:pi1} we know that $\pi\sps{0}\in K$ and $\pi\sps{1}\in K$.
Then by \eqref{eq:fpiineq}, 
\eq{\label{eq:meanvalue}
  \begin{aligned}
  \norm{\pi\sps{2}-\pi\sps{1}}&\leq \frac{1}{\omega}\norm{f(\pi\sps{1})-f(\pi\sps{0})},\\
  &=\frac{1}{\omega}\norm{\grad{f}(\pi\sps{0}+\zeta((\pi\sps{1}-\pi\sps{0})))(\pi\sps{1}-\pi\sps{0})}\\
  &=\frac{1}{\omega}\norm{(\grad{f}(\pi\sps{0}+\zeta((\pi\sps{1}-\pi\sps{0})))-\grad{f}(\pi^*))(\pi\sps{1}-\pi\sps{0})}\\
  &\leq\frac{M}{\omega}\norm{(\pi\sps{0}+\zeta((\pi\sps{1}-\pi\sps{0}))-\pi^*)(\pi\sps{1}-\pi\sps{0})}\\
  &\leq\frac{M}{\omega}\max\{\norm{\pi\sps{1}-\pi^*}, \norm{\pi\sps{0}-\pi^*}\}\norm{\pi\sps{1}-\pi\sps{0}}
  \end{aligned}
}
where we have used the identity $\grad f(\pi^*)(\pi\sps{1}-\pi\sps{0})=0$ and the fact that $\pi\sps{1}$ and $\pi\sps{0}$ are contained in $K$. In fact, we can prove that 
\[\grad f(\pi^*)(\pi\sps{k+1}-\pi\sps{k})=0, \quad \text{for any } k,\]
as follows. Since $f(\pi)_{sa} = -(r_s^a-((I-\gamma P^a)v_\pi)_s)$ has a similar form with $E(\pi)$, we can directly obtain $\grad f(\pi)$
\eq{
  (\grad f(\pi))_{sa, tb} = \lambda_{sa, t}(\pi) \left(-f(\pi)_{tb}+\tilde{c}(\pi)_t-\grad\Phi(\pi)_{tb}\right),
}
where $\lambda_{sa,t}(\pi) = Z_\pi^{-\top}\tilde{\rho}_{sa}$ and $\tilde{\rho}_{sa}$ is the $s$-th row of $I-\gamma P^a$. Since $f(\pi^*)+\grad\Phi(\pi^*) = B^\top c(\pi^*)$, we have
\begin{equation}
\label{eq:KKT}
\begin{aligned}
    &(\grad f(\pi^*)(\pi\sps{k+1}-\pi\sps{k}))_{sa}\\
    =& \sum_{tb} \lambda_{sa, t}(\pi^*) \left(-f(\pi^*)_{tb}+\tilde{c}(\pi^*)_t-\grad\Phi(\pi^*)_{tb}\right)(\pi_{tb}\sps{k+1}-\pi_{tb}\sps{k})\\
    =& \sum_{tb} \lambda_{sa, t}(\pi^*) \left(\tilde{c}(\pi^*)_t-c(\pi^*)_t\right)(\pi_{tb}\sps{k+1}-\pi_{tb}\sps{k})\\
    =& \sum_{t} \left[\Bigg(\lambda_{sa, t}(\pi^*) \left(\tilde{c}(\pi^*)_t-c(\pi^*)_t\right)\Bigg)\left(\sum_b(\pi_{tb}\sps{k+1}-\pi_{tb}\sps{k})\right)\right]\\
    =&~0,
\end{aligned}
\end{equation}
where the last equality results from the fact that $\sum_b\pi_{tb}\sps{k+1}=\sum_b\pi_{tb}\sps{k}=1$ for any $t$. Now from\eqref{eq:pi0}, \eqref{eq:pi1} and \eqref{eq:meanvalue}, we obtain
\eq{
\begin{aligned}
  \norm{\pi\sps{2}-\pi\sps{1}}&\leq\frac{M}{\omega}\max\{\norm{\pi\sps{1}-\pi^*}, \norm{\pi\sps{0}-\pi^*}\}\norm{\pi\sps{1}-\pi\sps{0}}\\
  &\leq \frac{M}{\omega}\cdot\frac{1}{16}\min\{\frac{\omega}{M}, \delta_0\}\norm{\pi\sps{1}-\pi\sps{0}}\\
  &\leq (\frac{1}{2}-\frac{1}{2^{1+2}})\norm{\pi\sps{1}-\pi\sps{0}}
\end{aligned}
}
Now, assuming that the induction hypothesis \eqref{eq:induce} holds for some $n\geq1$, we have
\eq{\label{eq:indnext1}
\begin{aligned}
  \norm{\pi\sps{n+1}-\pi\sps{*}}&\leq\norm{\pi\sps{n+1}-\pi\sps{n}}+\norm{\pi\sps{n}-\pi^*}\\
  &\leq \left(\prod_{k=1}^{n}(\frac{1}{2}-\frac{1}{2^{k+2}})\right)\norm{\pi\sps{1}-\pi\sps{0}}+\norm{\pi\sps{n}-\pi^*}\\
  &\leq \frac{1}{2^n}\cdot \frac{1}{8}\min\{\frac{\omega}{M}, \delta_0\}+(\frac{1}{2}-\frac{1}{2^{n+2}})\min\{\frac{\omega}{M}, \delta_0\}\\
  &=(\frac{1}{2}-\frac{1}{2^{n+2}}+\frac{1}{2^{n+3}})\min\{\frac{\omega}{M}, \delta_0\}\\
  &=(\frac{1}{2}-\frac{1}{2^{(n+1)+2}})\min\{\frac{\omega}{M}, \delta_0\},\\
\end{aligned}
}
which also implies that $\pi\sps{n+1}\in K$. Now using the same reasoning as \eqref{eq:meanvalue} but $(\pi\sps{0}, \pi\sps{1}, \pi\sps{2})$ replaced by $(\pi\sps{n}, \pi\sps{n+1}, \pi\sps{n+2})$, one obtains
\eq{
   \norm{\pi\sps{n+2}-\pi\sps{n+1}}\leq \frac{M}{\omega}\max\{\norm{\pi\sps{n+1}-\pi^*}, \norm{\pi\sps{n}-\pi^*}\}\norm{\pi\sps{n+1}-\pi\sps{n}}.
}
After plugging \eqref{eq:indnext1} and the induction hypothesis into this inequality, we get
\eq{\label{eq:indnext2}
\begin{aligned}
  \norm{\pi\sps{n+2}-\pi\sps{n+1}}&\leq \frac{M}{\omega}\cdot (\frac{1}{2}-\frac{1}{2^{n+3}})\min\{\frac{\omega}{M}, \delta_0\}\norm{\pi\sps{n+1}-\pi\sps{n}}\\
  &\leq(\frac{1}{2}-\frac{1}{2^{(n+1)+2}})\norm{\pi\sps{n+1}-\pi\sps{n}}
  \end{aligned}
}
With \eqref{eq:indnext1} and \eqref{eq:indnext2} we have shown that \eqref{eq:induce} holds with $n$ replaced by $n+1$. As a result, \eqref{eq:induce} holds for any $n$. From the second inequality in \eqref{eq:induce}, it is clear that $\pi\sps{k}$ converges (at least exponentially fast). Denote the limit of $\pi\sps{k}$ by $\tilde{\pi}$ for now, we obtain from \eqref{eq:replace} that
\eq{
  f(\tilde{\pi}) + \grad\Phi(\tilde{\pi}) - B^\top c(\tilde{\pi}) = 0,
}
thus $\tilde{\pi}=\pi^*$ by \Cref{thm:flowconv}.

  \textbf{Step 3. Prove the convergence rate is quadratic. } 
  Since $\pi\sps{k}$ converges to $\pi^*$ and $\grad f$ is Lipschitz continuous on $K$, we have
\eq{\label{eq:Lip}
    \underset{k\rightarrow\infty}{\lim}\frac{f(\pi\sps{k+1}) - f(\pi\sps{k}) - \grad f(\pi^*)\left(\pi\sps{k+1}-\pi\sps{k}\right)}{\norm{\pi\sps{k+1}-\pi\sps{k}}}=0.
}
On the other hand, we have
\begin{equation}
    \begin{aligned}
        &f(\pi\sps{k+1}) - f(\pi\sps{k}) - \grad f(\pi^*)\left(\pi\sps{k+1}-\pi\sps{k}\right) \\
  =&f(\pi\sps{k+1})+\grad\Phi(\pi\sps{k+1})-B^\top c(\pi\sps{k})-\grad f(\pi^*)\left(\pi\sps{k+1}-\pi\sps{k}\right)\\
  =&f(\pi\sps{k+1})+\grad\Phi(\pi\sps{k+1})-B^\top c(\pi\sps{k}),
    \end{aligned}
\end{equation}
where we have used \eqref{eq:KKT}. Combining with \eqref{eq:Lip} we arrive at
\eq{\label{eq:supconvergepre}
    \underset{k\rightarrow\infty}{\lim}\frac{f(\pi\sps{k+1})+\grad\Phi(\pi\sps{k+1})-B^\top c(\pi\sps{k})}{\norm{\pi\sps{k+1}-\pi\sps{k}}}=0.
}
With the last three lines of \eqref{eq:updateineq}, we obtain
\eq{
    \underset{k\rightarrow\infty}{\lim}\frac{\left(\grad\Phi(\pi\sps{k+2})-\grad\Phi(\pi\sps{k+1})\right)^\top(\pi\sps{k+2}-\pi\sps{k+1})}{\norm{\pi\sps{k+1}-\pi\sps{k}}\norm{\pi\sps{k+2}-\pi\sps{k+1}}}=0,
}
by multiplying the unit vector $\frac{\pi\sps{k+2}-\pi\sps{k+1}}{\norm{\pi\sps{k+2}-\pi\sps{k+1}}}$ to the fraction in \eqref{eq:supconvergepre}. Then by \eqref{eq:strong} we get
\eq{
    0=\underset{k\rightarrow\infty}{\lim}\frac{\norm{\pi\sps{k+2}-\pi\sps{k+1}}^2}{\norm{\pi\sps{k+1}-\pi\sps{k}}\norm{\pi\sps{k+2}-\pi\sps{k+1}}}= \underset{k\rightarrow\infty}{\lim}\frac{\norm{\pi\sps{k+2}-\pi\sps{k+1}}}{\norm{\pi\sps{k+1}-\pi\sps{k}}},
}
from which we can conclude that $\pi\sps{k}$ converges to $\pi^*$ superlinearly, i.e., 
\begin{equation}\label{eq:suplin}
\underset{k\rightarrow\infty}{\lim}\frac{\|\pi\sps{k+1}-\pi^*\|}{\|\pi\sps{k}-\pi^*\|} = 0.
\end{equation}
In fact, for any $\epsilon$ (assume $\epsilon<1/2$ without loss of generality), there is some $k(\epsilon)$ such that for any $k>k(\epsilon)$, $\frac{\norm{\pi\sps{k+2}-\pi\sps{k+1}}}{\norm{\pi\sps{k+1}-\pi\sps{k}}} < \epsilon$, then for any $k>k(\epsilon)$
\eqs{
    \norm{\pi\sps{k+1}-\pi^*}&\leq \sum_{n=k+1}^\infty\norm{\pi\sps{n+1}-\pi\sps{n}}\leq \sum_{n=k+1}^\infty\epsilon^{n-k}\norm{\pi\sps{k+1}-\pi\sps{k}}\\
    &\leq \frac{\epsilon}{1-\epsilon}\norm{\pi\sps{k+1}-\pi\sps{k}}\leq 2\epsilon \norm{\pi\sps{k+1}-\pi\sps{k}}.
}
Then
\eqs{
    \norm{\pi\sps{k}-\pi^*} &\geq\norm{\pi\sps{k+1}-\pi\sps{k}}-\norm{\pi\sps{k+1}-\pi^*}\\
    &\geq (\frac{1}{2\epsilon}-1)\norm{\pi\sps{k+1}-\pi^*}.
}
For any $G>0$, take $\epsilon = 1/(2G+2)$, then for any $k>k(\epsilon)$,
\eq{
    \norm{\pi\sps{k}-\pi^*} \geq (\frac{1}{2\epsilon}-1)\norm{\pi\sps{k+1}-\pi^*} = (G+1)\norm{\pi\sps{k+1}-\pi^*}>G\norm{\pi\sps{k+1}-\pi^*},
}
which shows that $\underset{k\rightarrow\infty}{\lim}\frac{\|\pi\sps{k}-\pi^*\|}{\|\pi\sps{k+1}-\pi^*\|} = +\infty$ and thus \eqref{eq:suplin} holds. Now, from \eqref{eq:replace} and \eqref{eq:KKT} we have
\eqs{
&\norm{f(\pi\sps{k+1})+\grad\Phi(\pi\sps{k+1})-B^\top c(\pi\sps{k})}\\
=&\norm{f(\pi\sps{k+1}) - f(\pi\sps{k}) - \grad f(\pi^*)\left(\pi\sps{k+1}-\pi\sps{k}\right)}\\
=&\norm{\left(\int_0^1\left[\grad f(\pi\sps{k}+t(\pi\sps{k+1}-\pi\sps{k}))-\grad f(\pi^*)\right]\mathrm{d} t\right) \left(\pi\sps{k+1}-\pi\sps{k}\right)}\\
\leq&~\tilde{C}\norm{\pi\sps{k}-\pi^*}\norm{\pi\sps{k+1}-\pi\sps{k}}
}
for some constant $\tilde{C}$, where we used \eqref{eq:suplin} and the Lipschitz continuity of $\grad f$ in the last equality. Multiplying both sides by ${\pi\sps{k+2}-\pi\sps{k+1}}$, and by \eqref{eq:strong} and the last three lines of \eqref{eq:updateineq} we have
\eqs{
    &\omega\norm{\pi\sps{k+2}-\pi\sps{k+1}}^2\\
    \leq&\left(\grad\Phi(\pi\sps{k+2})-\grad\Phi(\pi\sps{k+1})\right)^\top(\pi\sps{k+2}-\pi\sps{k+1})\\
    =&\left(f(\pi\sps{k+1})+\grad\Phi(\pi\sps{k+1})-B^\top c(\pi\sps{k})\right)^\top(\pi\sps{k+2}-\pi\sps{k+1})\\
    \leq&~\tilde{C}\norm{\pi\sps{k}-\pi^*}\norm{\pi\sps{k+1}-\pi\sps{k}}\norm{\pi\sps{k+2}-\pi\sps{k+1}},
}
which implies that
\eq{\label{eq:quadpre}\norm{\pi\sps{k+2}-\pi\sps{k+1}} \leq \tilde{C}\norm{\pi\sps{k}-\pi^*}\norm{\pi\sps{k+1}-\pi\sps{k}},}
with some constant $\tilde{C}$. From \eqref{eq:suplin}, we have
\eq{
    \underset{k\rightarrow\infty}{\lim}\frac{\|\pi\sps{k}-\pi\sps{k+1}\|}{\|\pi\sps{k}-\pi^*\|} = \underset{k\rightarrow\infty}{\lim}\frac{\|\pi\sps{k+1}-\pi\sps{k+2}\|}{\|\pi\sps{k+1}-\pi^*\|} = 1.
}
Combining this with \eqref{eq:quadpre} leads to
\eq{
    \norm{\pi\sps{k+1}-\pi^*} \leq C\norm{\pi\sps{k}-\pi^*}^2,
}
for some constant $C$, which closes the proof.
\end{proof}

\section{Conclusion and Discussion}\label{sec:conclusion}

In this paper, we present a fast approximate Newton method for the policy gradient algorithm with provable quadratic convergence. The proposed method gives a systematic theory that includes the well-known natural policy gradient algorithm as a special case and naturally extends to other regularizers such as the reverse KL divergence, the Hellinger divergence, and the $\alpha$-divergence.

With a relatively simple proof, we show the local quadratic convergence of the proposed approximate Newton method as well as the global convergence of the approximate Newton gradient flow to the optimal solution. The quadratic convergence is confirmed numerically on both medium and
large sparse models. In contrast with mirror descent type first-order methods (e.g. \cite{zhan2021policy}) that take up to tens of thousands of iterations even with manually tuned learning rate, the proposed approximate Newton algorithms typically converge in less than $10$ iterations, despite the large discount rate ($\approx 1$) and small regularization coefficient ($\approx 0$). 

For future work, we plan to adapt the technique used here to other gradient-based algorithms for
solving the MDP problems. Other forms of $f$-divergence can also be included. An interesting direction is to apply different types of numerical schemes for ordinary differential equations (ODEs) to the approximate Newton gradient flow presented in \Cref{sec:flowconv}, which can be helpful for obtaining a good initial policy such that the discrete approximate Newton method is able to achieve fast quadratic convergence. Another direction is
to consider continuous MDP problems by leveraging function approximation, effective spatial
discretization, or model reduction.

\bibliographystyle{siamplain}
\bibliography{references}
\end{document}

%% file: ex_shared.tex
\usepackage{amsmath,amssymb,a4wide}
\usepackage{latexsym}
\usepackage{indentfirst}
\usepackage{graphicx}
\usepackage{placeins}
\usepackage{booktabs}
\usepackage{color}
\usepackage{algorithm}
\usepackage{algorithmic}
\usepackage{multirow}
\usepackage{enumerate}
\usepackage{subfigure}

\usepackage{lipsum}
\usepackage{amsfonts}
\usepackage{graphicx}
\usepackage{epstopdf}
\usepackage{algorithmic}
\ifpdf
  \DeclareGraphicsExtensions{.eps,.pdf,.png,.jpg}
\else
  \DeclareGraphicsExtensions{.eps}
\fi

\newsiamremark{remark}{Remark}
\newsiamremark{hypothesis}{Hypothesis}
\crefname{hypothesis}{Hypothesis}{Hypotheses}
\newsiamthm{claim}{Claim}
\newsiamthm{prop}{Proposition}

\headers{Approximate Newton policy gradient algorithms}{H. Li, S. Gupta, H. Yu, L. Ying, and I. Dhillon}

\title{Approximate Newton policy gradient algorithms\thanks{Accepted by SIAM SISC May 19th, 2023.
}}

\author{Haoya Li\thanks{Stanford University, Stanford, CA
  (\email{lihaoya@stanford.edu}).}
\and Samarth Gupta\thanks{Carnegie Mellon University, Pittsburgh, PA
  (\email{samarthg@andrew.cmu.edu}).}
\and Hsiangfu Yu\thanks{Amazon Search, Palo Alto, CA
  (\email{rofu.yu@gmail.com}).}
\and Lexing Ying\thanks{Stanford University, Stanford, CA; Amazon Search, Palo Alto, CA
  (\email{lexing@gmail.com}).}
\and Inderjit Dhillon\thanks{University of Texas at Austin and Google, work done while at Amazon Search
  (\email{inderjit@cs.utexas.edu}).}}

\usepackage{amsopn}

\newcommand{\sps}[1]{^{(#1)}}

\newcommand{\diag}{\mathrm{diag}}

\newcommand{\norm}[1]{\left\lVert#1\right\rVert}

\newcommand{\eq}[1]{\begin{align}#1\end{align}}
\newcommand{\eqs}[1]{\begin{align*}#1\end{align*}}
\newcommand{\bbm}{\begin{bmatrix}}
\newcommand{\ebm}{\end{bmatrix}}

\newcommand{\E}{\mathbb{E}}
\newcommand{\R}{\mathbb{R}}

\newcommand{\grad}{\nabla}

\newcommand{\g}{\gamma}


%% file: ex_article.bbl
\begin{thebibliography}{10}

\bibitem{agarwal2020optimality}
{\sc A.~Agarwal, S.~M. Kakade, J.~D. Lee, and G.~Mahajan}, {\em Optimality and
  approximation with policy gradient methods in {M}arkov decision processes},
  in {Conference on Learning Theory}, PMLR, 2020.

\bibitem{ahmed2019understanding}
{\sc Z.~Ahmed, N.~Le~Roux, M.~Norouzi, and D.~Schuurmans}, {\em Understanding
  the impact of entropy on policy optimization}, in {International Conference
  on Machine Learning}, PMLR, 2019.

\bibitem{ali1966general}
{\sc S.~M. Ali and S.~D. Silvey}, {\em A general class of coefficients of
  divergence of one distribution from another}, Journal of the Royal
  Statistical Society: Series B (Methodological), 28 (1966), pp.~131--142.

\bibitem{bellman1957markovian}
{\sc R.~Bellman}, {\em A markovian decision process}, Journal of mathematics
  and mechanics, 6 (1957), pp.~679--684.

\bibitem{boyd2004convex}
{\sc S.~Boyd, S.~P. Boyd, and L.~Vandenberghe}, {\em Convex optimization},
  Cambridge university press, 2004.

\bibitem{cen2020fast}
{\sc S.~Cen, C.~Cheng, Y.~Chen, Y.~Wei, and Y.~Chi}, {\em Fast global
  convergence of natural policy gradient methods with entropy regularization},
  July 2020, \url{https://arxiv.org/abs/2007.06558}.

\bibitem{de1998comparison}
{\sc L.~G. de~Pillis}, {\em A comparison of iterative methods for solving
  nonsymmetric linear systems}, Acta Applicandae Mathematica, 51 (1998),
  pp.~141--159.

\bibitem{dennis1974characterization}
{\sc J.~E. Dennis and J.~J. Mor{\'e}}, {\em A characterization of superlinear
  convergence and its application to quasi-newton methods}, Mathematics of
  computation, 28 (1974), pp.~549--560.

\bibitem{geist2019theory}
{\sc M.~Geist, B.~Scherrer, and O.~Pietquin}, {\em A theory of regularized
  {M}arkov decision processes}, in {International Conference on Machine
  Learning}, PMLR, 2019.

\bibitem{gunasekar2021mirrorless}
{\sc S.~Gunasekar, B.~Woodworth, and N.~Srebro}, {\em Mirrorless mirror
  descent: A natural derivation of mirror descent}, in International Conference
  on Artificial Intelligence and Statistics, 2021.

\bibitem{hiriart2004fundamentals}
{\sc J.-B. Hiriart-Urruty and C.~Lemar{\'e}chal}, {\em Fundamentals of convex
  analysis}, Springer Science \& Business Media, 2004.

\bibitem{kakade2001natural}
{\sc S.~M. Kakade}, {\em A natural policy gradient}, in {Advances in Neural
  Information Processing Systems}, 2001.

\bibitem{konda2000actor}
{\sc V.~R. Konda and J.~N. Tsitsiklis}, {\em Actor-critic algorithms}, in
  {Advances in Neural Information Processing Systems}, 2000.

\bibitem{lan2021policy}
{\sc G.~Lan}, {\em Policy mirror descent for reinforcement learning: Linear
  convergence, new sampling complexity, and generalized problem classes}, Feb.
  2021, \url{https://arxiv.org/abs/2102.00135}.

\bibitem{li2021softmax}
{\sc G.~Li, Y.~Wei, Y.~Chi, Y.~Gu, and Y.~Chen}, {\em Softmax policy gradient
  methods can take exponential time to converge}, Feb. 2021,
  \url{https://arxiv.org/abs/2102.11270}.

\bibitem{martens2020new}
{\sc J.~Martens}, {\em New insights and perspectives on the natural gradient
  method}, Journal of Machine Learning Research, 21 (2020), pp.~1--76.

\bibitem{mei2020global}
{\sc J.~Mei, C.~Xiao, C.~Szepesvari, and D.~Schuurmans}, {\em On the global
  convergence rates of softmax policy gradient methods}, in {International
  Conference on Machine Learning}, PMLR, 2020.

\bibitem{mensch2018differentiable}
{\sc A.~Mensch and M.~Blondel}, {\em Differentiable dynamic programming for
  structured prediction and attention}, in International Conference on Machine
  Learning, 2018.

\bibitem{mnih2016asynchronous}
{\sc V.~Mnih, A.~P. Badia, M.~Mirza, A.~Graves, T.~Lillicrap, T.~Harley,
  D.~Silver, and K.~Kavukcuoglu}, {\em Asynchronous methods for deep
  reinforcement learning}, in {International Conference on Machine Learning},
  PMLR, 2016.

\bibitem{nemirovskij1983problem}
{\sc A.~S. Nemirovskij and D.~B. Yudin}, {\em Problem complexity and method
  efficiency in optimization}, Wiley, 1983.

\bibitem{nesterov2009primal}
{\sc Y.~Nesterov}, {\em Primal-dual subgradient methods for convex problems},
  Mathematical programming, 120 (2009), pp.~221--259.

\bibitem{neu2017unified}
{\sc G.~Neu, A.~Jonsson, and V.~G{\'o}mez}, {\em A unified view of
  entropy-regularized {M}arkov decision processes}, May 2017,
  \url{https://arxiv.org/abs/1705.07798}.

\bibitem{raskutti2015information}
{\sc G.~Raskutti and S.~Mukherjee}, {\em The information geometry of mirror
  descent}, IEEE Transactions on Information Theory, 61 (2015), pp.~1451--1457.

\bibitem{renyi1961measures}
{\sc A.~R{\'e}nyi}, {\em On measures of entropy and information}, in
  {Proceedings of the Fourth Berkeley Symposium on Mathematical Statistics and
  Probability, Volume 1: Contributions to the Theory of Statistics}, 1961.

\bibitem{rodomanov2021new}
{\sc A.~Rodomanov and Y.~Nesterov}, {\em New results on superlinear convergence
  of classical quasi-newton methods}, Journal of optimization theory and
  applications, 188 (2021), pp.~744--769.

\bibitem{rodomanov2021rates}
{\sc A.~Rodomanov and Y.~Nesterov}, {\em Rates of superlinear convergence for
  classical quasi-newton methods}, Mathematical Programming,  (2021),
  pp.~1--32.

\bibitem{saad2003iterative}
{\sc Y.~Saad}, {\em Iterative methods for sparse linear systems}, SIAM, 2003.

\bibitem{schulman2015trust}
{\sc J.~Schulman, S.~Levine, P.~Abbeel, M.~Jordan, and P.~Moritz}, {\em Trust
  region policy optimization}, in {International Conference on Machine
  Learning}, PMLR, 2015.

\bibitem{schulman2015high}
{\sc J.~Schulman, P.~Moritz, S.~Levine, M.~Jordan, and P.~Abbeel}, {\em
  High-dimensional continuous control using generalized advantage estimation},
  June 2015, \url{https://arxiv.org/abs/1506.02438}.

\bibitem{schulman2017proximal}
{\sc J.~Schulman, F.~Wolski, P.~Dhariwal, A.~Radford, and O.~Klimov}, {\em
  Proximal policy optimization algorithms}, July 2017,
  \url{https://arxiv.org/abs/1707.06347}.

\bibitem{shani2020adaptive}
{\sc L.~Shani, Y.~Efroni, and S.~Mannor}, {\em Adaptive trust region policy
  optimization: Global convergence and faster rates for regularized mdps}, in
  {Proceedings of the AAAI Conference on Artificial Intelligence}, 2020.

\bibitem{silver2014deterministic}
{\sc D.~Silver, G.~Lever, N.~Heess, T.~Degris, D.~Wierstra, and M.~Riedmiller},
  {\em Deterministic policy gradient algorithms}, in {International Conference
  on Machine Learning}, PMLR, 2014.

\bibitem{sutton2018reinforcement}
{\sc R.~S. Sutton and A.~G. Barto}, {\em Reinforcement learning: An
  introduction}, MIT press, 2018.

\bibitem{sutton1999policy}
{\sc R.~S. Sutton, D.~A. McAllester, S.~P. Singh, and Y.~Mansour}, {\em Policy
  gradient methods for reinforcement learning with function approximation}, in
  {Advances in Neural Information Processing Systems}, 2000.

\bibitem{tomar2020mirror}
{\sc M.~Tomar, L.~Shani, Y.~Efroni, and M.~Ghavamzadeh}, {\em Mirror descent
  policy optimization}, May 2020, \url{https://arxiv.org/abs/2005.09814}.

\bibitem{van1992bi}
{\sc H.~A. Van~der Vorst}, {\em Bi-cgstab: A fast and smoothly converging
  variant of bi-cg for the solution of nonsymmetric linear systems}, SIAM
  Journal on scientific and Statistical Computing, 13 (1992), pp.~631--644.

\bibitem{wang2021hessian}
{\sc L.~Wang and M.~Yan}, {\em Hessian informed mirror descent}, June 2021,
  \url{https://arxiv.org/abs/2106.13477}.

\bibitem{williams1992simple}
{\sc R.~J. Williams}, {\em Simple statistical gradient-following algorithms for
  connectionist reinforcement learning}, Machine learning, 8 (1992),
  pp.~229--256.

\bibitem{ying2020mirror}
{\sc L.~Ying}, {\em Mirror descent algorithms for minimizing interacting free
  energy}, Journal of Scientific Computing, 84 (2020), pp.~1--14.

\bibitem{zhan2021policy}
{\sc W.~Zhan, S.~Cen, B.~Huang, Y.~Chen, J.~D. Lee, and Y.~Chi}, {\em Policy
  mirror descent for regularized reinforcement learning: A generalized
  framework with linear convergence}, May 2021,
  \url{https://arxiv.org/abs/2105.11066}.

\end{thebibliography}
